\documentclass[10pt,twocolumn]{article} 

\usepackage{simpleConference}
\usepackage{times}
\usepackage{graphicx}
\usepackage{amssymb}
\usepackage{latexsym}

\usepackage{tikz}
\usepackage{booktabs}
\usepackage{multirow}
\usepackage{comment}
\usepackage{subcaption}
\usepackage{amsmath,amssymb}
\usepackage{graphicx}

\usepackage{graphbox,xcolor}
\usepackage[pagebackref,breaklinks,colorlinks]{hyperref}

\begin{document}

\title{Zero-Shot Video Captioning with Evolving Pseudo-Tokens}

\author{Yoad Tewel \quad\quad\quad Yoav Shalev\quad\quad\quad Roy Nadler\quad\quad\quad Idan Schwartz \quad\quad\quad Lior Wolf\\
School of Computer Science, Tel Aviv University
}

\maketitle
\thispagestyle{empty}

\begin{abstract}
We introduce a zero-shot video captioning method that employs two frozen networks: the GPT-2 language model and the CLIP image-text matching model. The matching score is used to steer the language model toward generating a sentence that has a high average matching score to a subset of the video frames. Unlike zero-shot image captioning methods, our work considers the entire sentence at once. This is achieved by optimizing, during the generation process, part of the prompt from scratch, by modifying the representation of all other tokens in the prompt, and by repeating the process iteratively, gradually improving the specificity and comprehensiveness of the generated sentence. Our experiments show that the generated captions are coherent and display a broad range of real-world knowledge. Our code is available at: \url{https://github.com/YoadTew/zero-shot-video-to-text}.
\end{abstract}

\section{Introduction}
Image captioning is becoming increasingly accurate and can successfully tackle more complex benchmarks than ever before. However, the progress in video captioning is slower due to both methodological reasons and dataset construction challenges. First, the video captioning task itself has more possible definitions than image captioning. For example, do we want a complete description of the events in the video or a general description of it? Second, even considerably more limited tasks, such as action recognition in pre-trimmed videos, are still technologically challenging. Third, the descriptions attached to web videos are often not an accurate depiction of the content and events in the video, making the construction of web-scale datasets more challenging.

These challenges mean that strategies used in related tasks are less suitable for the task of video captioning. (i) One cannot train on large and noisy datasets, since the amount of noise would be too high. (ii) Learning on a carefully curated dataset would be too restrictive in terms of the obtained coverage, and too limited in the use of language (due to the nature of cost-effective human annotation). (iii) A divide-and-conquer approach that captions single frames would still require a sophisticated way of combining all information and text. (iv) Relying on pretrained action recognition models would lead to inaccurate results that are very much limited in the scope of captured actions.

In light of the challenges faced by model-learning approaches, the method we introduce is a zero-shot method. It uses the information stored in two pre-trained and frozen networks to perform the video captioning task. One model is an autoregressive language model that can generate natural and mostly logical sentences %one word at a time. 
The second model is an image-text matching model that is used to steer the language model, via a contrastive loss, toward sentences that match a set of input frames. 

\begin{figure*}[t]
	\centering
    \includegraphics[width=0.9\linewidth]{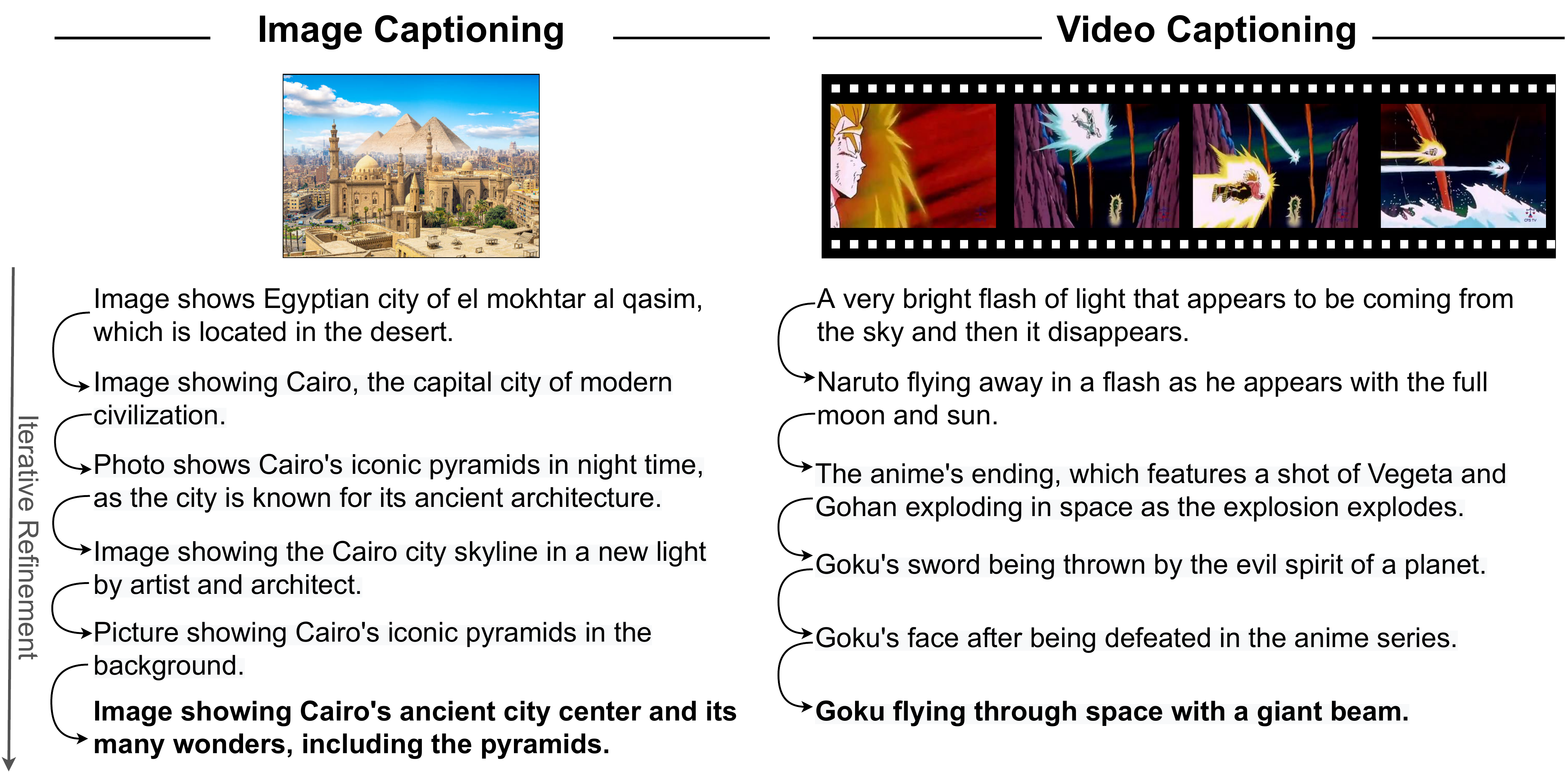}
    \caption{We present a novel way of optimizing a sentence generation process to match a set of images by using pseudo-tokens and iteratively generating sentences. The method exhibits real-world knowledge, which enables it to construct a narrative from images and describe videos. This narrative evolves through the iterations and tends to become increasingly specific.} 
    \label{fig:progression sample}
    
\end{figure*}

We make extensive use of the fact that the autoregressive process is conditioned on %both selecting the next word that leads to a better match and altering 
its prompt. Our method's generative process employs a prompt that contains three parts: (i) pseudo-tokens that are vectors in the latent space of the language models, (ii) a random prompt such as ``Image of'' that provides context for the captioning task, but also varies (``Photo of'', ``Video of'', etc.) as a form of inference-time augmentation, and (iii) the previously generated tokens. 

The first part differs from the other two in that it does not originate from actual language tokens. However, all three parts are manipulated, in memory, by optimizing them to increase the image-matching score of the next word being selected. This is done while regularizing in such a way that the language model does not drastically alter the likelihood of the next word.

Since the sentence generation process is autoregressive, the initial words of the generated sentence employed psuedo-tokens that were not optimized based on the signal obtained from the entire sentence. We, therefore, repeat the process and start the autoregressive process with the pseudo-tokens that were obtained at the end of the previous generation iteration. As we show in Fig.~\ref{fig:progression sample}, this leads to increasingly concrete prompts. The process is repeated sixteen times, and the sentence that maximizes the image-text matching score is selected.

In our implementation, we use the GPT-2 language model~\cite{radford2019language}, due to its availability, and the CLIP image-text matching model~\cite{radford2021learning}, which is often used in zero-shot learning. We experiment with both video captioning and describing image sets. Our results show a clear advantage over the state-of-the-art video captioning methods and over recent zero-shot image captioning methods. Our code is attached as supplementary.

\section{Related Work}
Image captioning is a fundamental vision and language task. Early methods applied RNNs~\cite{Mao2014DeepCW,klein2014fisher}. Attention was added to identify relevant salient objects~\cite{xu2015show,schwartz2017high}. Graph neural networks and transformers helped model spatial and semantic interactions~\cite{yao2018exploring,kipf2016semi,yang2019auto}. %,vaswani2017attention,pan2020x,dosovitskiy2021image,guo2020normalized,schwartz2019factor}. 

Other video-based tasks include action recognition~\cite{SimonyanNIPS2014}, paragraph captioning~\cite{yu2016video}, and video object segmentation~\cite{Perazzi2016}. %Different methods have been studied for encoding spatio-temporal information. C3D employs 3D CNN~\cite{tran2015learning}. I3D is trained on a large action recognition dataset~\cite{carreira2017quo}. 
Some approaches to video analysis use different experts, including speech, audio, motion, OCR, appearance, and face detection~\cite{liu2019use,akbari2021vatt}. This work considers video captioning by generating a single sentence that adequately describes a set of frames. Despite the lack of temporal information in a set, we find that the language model generates a logical order of events.

 In various contributions, sparse sampling along with better spatial reasoning have proved sufficient for handling reasoning tasks, such as video dialogs~\cite{schwartz2019simple} and video retrieval~\cite{lei2022revealing, zhou2017temporal,santoro2017simple,lei2021less}.  An attention module that selects the relevant frames can reduce the temporal dimension~\cite{ali2022video,gabeur2020multi}. In our work, we also employ sparse sampling that utilizes distances in the CLIP embedding space to construct a set of the most relevant frames. 
 
 Significant improvements have been achieved by using large-scale unsupervised vision-language data sets with millions of image and text pairs~\cite{zhang2021vinvl,devlin2018bert,li2020oscar} and million of videos~\cite{miech2019howto100m,zhu2020actbert,luo2020univl}. The unsupervised data is used in a pre-training phase. Fine-tuning for a particular task is done in the final stage, using smaller datasets annotated by humans. Although this process has been shown to boost performance, it remains highly dependent on the human annotations. Our experiments show that captions learned based on human annotations are dull. Moreover, the same repetitive patterns are produced by significantly different baseline methods~\cite{Perez-Martin_2021_WACV,chen2020delving}.
 
 As a way to reduce dependence on supervision, CLIP has been using web-based learning to study language-image relationships~\cite{radford2021learning}. CLIP's strength lies in the amount of data it collects. CLIP is trained in a contrastive manner, using 400M image-text pairs from the web, unlike supervised methods, which typically use thousands of samples. The method learns a bimodal distance metric between an input image and a text sentence, which we refer to as the CLIP score. Matching videos to text also benefited from a contrastive approach~\cite{xu2021videoclip,li2021align}. However, video data can be challenging to collect. 
%  A retrieval engine based solely on CLIP has been developed~\cite{fang2021clip2video,gao2021clip2tv}. 
 In our case, we used CLIP to guide a language generator in a zero-shot manner.
  
 Contrastive learning is effective for zero-shot capabilities, achieving good results for several tasks, including image classification and action recognition. However, generative tasks based directly on the CLIP score are rare, since the score requires seeing both text and image. Instead, multiple contributions rely on CLIP's image and text encodings, which are known to improve performance in vision+language tasks~\cite{shen2021much}, especially in image captioning~\cite{mokady2021clipcap} and video captioning~\cite{tang2021clip4caption}. However, fine-tuning distorts the latent semantics of CLIP's encoder~\cite{tewel2021zero}. MAGIC employs CLIP scores to shift PLM logits towards image correspondence~\cite{2022arXiv220502655S}. Despite this, they fine-tunes the PLM on the text corpus of MS-COCO captions, so robustness is still compromised.  Alternatively, CLIP can be used as part of a loss term to guide generative processes to match language and text, for example, as a loss term for 3D mesh generation~\cite{text2mesh} or text-guided image generation~\cite{patashnik2021styleclip, chefer2021image}.   %Using CLIP, we also apply guidance loss to describe a new type of data, an image set. 
 
 Recently, it was suggested to use CLIP loss to guide a Pretrained Language Model (PLM) for image captioning~\cite{tewel2021zero}. Compared with ours, their method optimizes each generated token individually, aiming to obtain the token closest to the given image. In contrast, our method optimizes pseudo-tokens through iteratively generating sentences, aiming to steer the generation process of the entire sentence. Although their process is effective in describing one visual cue, it is challenged by the more difficult task of describing multiple images coherently. We demonstrate that the ability to manipulate an entire sentence without committing to a single generation path has beneficial effects. Optimizing an entire sentence also means not requiring sequence generation strategies such as beam search.  %Our method guide PLM by employing pseudo-tokens as well as prefix tokens.
 
 The literature on tuning prior knowledge within large-scale PLMs, such as GPT-2~\cite{radford2019language}, is growing rapidly. There are several on-going directions: (i) Fine-tuning, e.g., by using GANs~\cite{ziegler2019finetuning} or RL~\cite{yu2016seqgan}; (ii) Disentangling the latent representations~\cite{shen2017style}; (iii) Training a controllable LM with fixed control codes~\cite{keskar2019ctrl}; (iv) Trainable decoding~\cite{gu2017trainable}; (v) Decoding steering~\cite{Dathathri2020Plug}; and (vi) Prompt engineering~\cite{shin2020autoprompt}, as well as prompt learning~\cite{wallace2019universal,li2021prefix,gao2020making,liu2021gpt}. In this work, we present a novel PLM decoding approach that combines steering and prompt tuning by generating sentences iteratively and applying prompt tuning.

\begin{figure*}[t]
	\centering
    \includegraphics[width=0.85\linewidth]{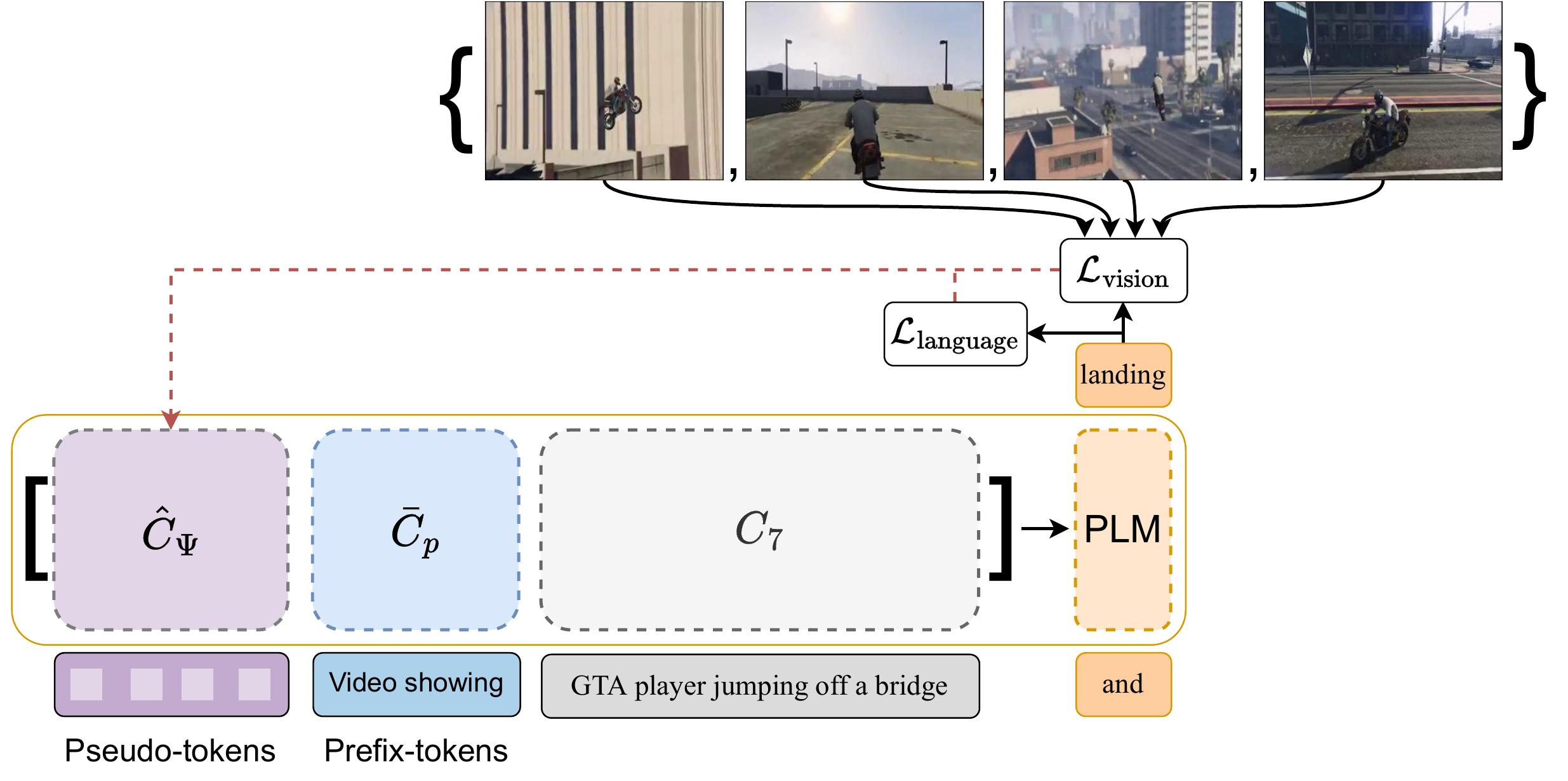}
    \caption{Illustration of our method for guiding a PLM to generate the word `landing'. Optimization takes place during the autoregression inference, by tuning pseudo-tokens ($\hat C_{\Psi}$).  Two signals steer the pseudo-tokens' representations, visual correspondence ($\mathcal{L}_{\text{vision}}$) and language fluency ($\mathcal{L}_{\text{language}}$).} 
    \label{fig:approach}
    
\end{figure*}

\section{Method}

Our goal is to create a sentence $S=\{t_1,\ldots, t_M\}$ of length $M$ that describes a set of video frames $\mathcal{F}=\{F_1,\ldots, F_N\}$, where $N$ is the number of frames. When $N=1$, the problem corresponds to traditional image captioning.

Two components are at the core of our solution. The first is a pre-trained language model (PLM) that generates sentences, for which we use GPT-2. The second, CLIP, is a pre-trained model that computes the distance between a frame $F$ and a sentence $S$, and guides the PLM during inference. % We employ the CLIP model, as a guidance during the auto-regressive process of the PLM. 

Guiding a PLM with CLIP has recently shown promising results for image captioning ~\cite{tewel2021zero, su2022language}.  These approaches use CLIP to optimize the next token to fit the image; we call this technique \textit{token-level optimization}. However, the application of these approaches to videos is limited, especially in the case of low homogeneity between frames. In this case, maintaining language fluency is difficult, since: (i) a single token has to describe a set of non-homogeneous frames, and (ii) the generation commits to a single direction, restricting the flexibility of the process. 
By contrast, rather than optimizing tokens, our method performs a \textit{sentence-level optimization}.   To achieve this, the inference starts with randomly initialized \textit{pseudo-tokens}. These tokens do not need to be actual words in the dictionary, but rather hidden states of words, that can be optimized using gradient descent. Description of visual content is driven using \textit{prefix-tokens}, such as `Video showing'. The next step consists of generating multiple sentences and continuously optimizing the pseudo-tokens. This is accomplished by calculating two types of losses: (i) $\mathcal{L}_{\text{vision}}$, which is the sum of the distance between all frames in $\mathcal{F}$ and the generated sentence, and (ii)  $\mathcal{L}_{\text{language}}$ which
takes into account language characteristics by considering the PLM token distribution. With no additional supervision or training, we benefit from the extensive knowledge embedded in CLIP and GPT-2. Our autoregressive process in depicted in Fig.~\ref{fig:approach}. %In the following section, we describe how we train pseudo prompt tokens to generate a description of the set of images.

\begin{table*}[t]
    \centering
	\resizebox{0.9\linewidth}{!}
	{
	    \begin{tabular}{llccccccccc} 
	    \toprule
	    && \multicolumn{5}{c}{\textbf{Supervised Metrics}}  & \multicolumn{4}{c}{\textbf{Unsupervised Metrics}}  \\ 	
		\cmidrule(lr){3-7} 	\cmidrule(lr){8-11}    
		Dataset & Method & B@4 & M & C & R & $\text{CLIP-S}^{\text{Ref}}$ & CLIP-S & BLIP-S & Retrieval & PP \\ 
		\midrule
		\multirow{5}{*}{MSR-VTT}&VNS-GRU~\cite{chen2020delving} & 0.453 & 0.299 & \bf 0.530 & 0.634 & 0.739 & 0.626 & 0.623 & 0.446 & 118.81 \\
		&SemSynAN~\cite{Perez-Martin_2021_WACV} & \bf 0.464 & \bf 0.304 & 0.519 & \bf 0.647 & 0.733 & 0.619 & 0.608 & 0.437 & 155.01 \\
	\cmidrule(lr){2-11} 
	&\multicolumn{9}{c}{\textit{Zero-Shot Methods}}\\
	\cmidrule(lr){2-11} 
	& ZeroCap*~\cite{tewel2021zero} & 0.023 & 0.129 & 0.058 & 0.304 & 0.739 & 0.710 & 0.575 & 0.442  & 54.71\\
		&MAGIC*~\cite{2022arXiv220502655S} & 0.055 & 0.133 & 0.074 & 0.354 & 0.628 & 0.566 & 0.434 & 0.392 & 30.48 \\
        & Ours & 0.030 & 0.146 & 0.113 & 0.277 & \bf 0.785 & \bf 0.775 & \bf 0.675 & \bf 0.504  & \bf 18.35\\
		\midrule
		\multirow{4}{*}{MSVD}&VNS-GRU~\cite{chen2020delving} & \bf 0.665 & \bf 0.421 & \bf 1.215 & \bf 0.797 & 0.780 & 0.673 & 0.646 & 0.557 & 418.72 \\
		&SemSynAN~\cite{Perez-Martin_2021_WACV} & 0.644 & 0.419 & 1.115 & 0.795 & 0.767 & 0.660 & 0.633 & 0.546 & 242.46 \\
	\cmidrule(lr){2-11} 
	&\multicolumn{9}{c}{\textit{Zero-Shot Methods}}\\
	\cmidrule(lr){2-11} 
		& ZeroCap*~\cite{tewel2021zero} & 0.029 & 0.163 & 0.096 & 0.354 & 0.762 & 0.765 & 0.642 & 0.500  & 28.44\\
		&MAGIC*~\cite{2022arXiv220502655S} & 0.066 & 0.161 & 0.140 & 0.401 & 0.670 & 0.623 & 0.497 & 0.469 & 29.84 \\
        & Ours & 0.030 & 0.178 & 0.174 & 0.314 & \bf 0.805 & \bf 0.822 & \bf 0.743 & \bf 0.569  & \bf 18.94\\
		\bottomrule
	    \end{tabular}
    }
    \medskip
	\caption{Quantitative results for video captioning. We separate the results into two categories: (i) supervised metrics that require human references,  B@4 = BLEU-4, M = METEOR, C = CIDEr, S = SPICE, and $\text{CLIP-S}^{\text{Ref}}$. (ii) Unsupervised metrics that use a pre-trained model, CLIP-S = CLIP-based image-text similarity, BLIP-S = BLIP-based image-text similarity~\cite{li2022blip}, Retrieval = VideoCLIP-based video-text similarity~\cite{xu2021videoclip}, and PP = caption perplexity computed with BERT~\cite{devlin2018bert}. (*) denotes that the model is adapted from image captioning to video captioning.% $\dag$ indicates a zero-shot model.
	}
	\label{tab:video-cap}
\end{table*}
\paragraph{PLM Guidance with Prompt Learning}

PLMs are trained on vast web knowledge to optimize a sum of conditionals, i.e.,
$\max_{\theta}p(S) = \sum_{i=1}^L p_{\theta}(w_i|w_{1:i-1}),$
where $\theta$ are trainable weights. The likelihood of each sub-sentence depends on its context. Thus, one can solve various tasks by altering the input context. For instance, to answer the question ``what is the capital of Britain?'' one could plug into the PLM the prompt ``The capital of Britain is.'' The PLM then finds the most likely next token (``London'') to optimize the conditional probability.

Prompt engineering entails finding the most suitable prompt for a given task. In our case, the task is to generate a sentence that maximizes similarity to a set of frames $\mathcal{F}$. Similarity is measured in terms of CLIP's distance metric between image and text. 
Any image imposes its own set of constraints, and the prompt needs to account for all of them. %Therefore, to optimize similarity with different images, we need to find the most appropriate prompt. 
The prompt must be flexible enough, which is ensured by optimizing pseudo-tokens, i.e., instead of finding real tokens for each video, we tune representative embeddings of tokens. 

The GPT-2 PLM is built with $L$ layers of Transformers, each composed of key and value embeddings, to model interactions between tokens \cite{vaswani2017attention}. The context of previous tokens can be cached by keeping their key and value representations. We denote the cache with $C_i=[K_j^l,V_j^l]_{j<i,l\le L}$,  where $i$ is the number of tokens, and $K_j^l,V_j^l$ are the key and value embeddings of the $l$-th Transformer layer of the $j$-th token. 

Our method starts the autoregression process with a randomly initialized cached pseudo-prompt context, i.e., $\hat C_{\Psi}=[\hat K_j^l,\hat V_j^l]_{0<j<\Psi,l\le L}$, which represents $\Psi$ pseudo-tokens. 

We further include $\bar C_p=[\bar K_j^l,\bar V_j^l]_{0<j<p, l\le L}$, where $p$ is the prefix length. The prefix-tokens serve to direct the task towards captioning a set of images. The prefix-tokens are sampled as one of the prompts in the set $\mathcal{P}= $\{``Image of'', ``Picture of'', ``Photo of'', ``Video of'',``Image shows'', ``Picture shows'', ``Photo shows'', ``Video shows'', ``Image showing'', ``Picture showing'', ``Photo showing'', ``Video showing''\}. 

Overall, the autoregressive process takes the form
\begin{align}
    \label{eq:plm_prompt}
    p_{i+1} (\hat C_{\Psi}) = \text{PLM}(t_{i}, [\hat C_{\Psi}, \bar C_p, C_i]),
\end{align}
where $p_{i+1}$ is the distribution of the next token. 

\paragraph{Loss:}   
At each step in the auto-regression process, we aggregate our loss, which will be used for optimization only after generating a complete sentence. Our first loss term encourages the generated text to correspond to the set of images.  

Let $S_k$ be the sentence generated up until this stage, ending with the token $k$. The visual-semantic loss calculates the cross-entropy ($\operatorname{CE}$) between the optimized PLM distribution and the CLIP potential similarity distribution $\theta_{\text{CLIP}}$:
\begin{align}
    \mathcal{L}_{\text{vision}}(\hat C_{\Psi})  =  \operatorname{CE} \left( p_{i+1}(\hat C_{\Psi}) , \theta_\text{CLIP} \right),
\end{align}
where $ \theta_{\text{CLIP}}(k) \propto \sum_{F\in \mathcal{F}} \operatorname{CLIP}\left( F,S_k\right)$ is the sum of CLIP's matching scores of $S_k$ with all the frames in $\mathcal{F}$. We compute the score for the top 100 tokens according to the original PLM distribution, with the rest set to zero.

While the PLM is trained on natural text, the model in which the free-form context $C_{\Psi}$ is added (Eq.~\ref{eq:plm_prompt}) can shift to very different distributions during optimization. In order to maintain fluent language, we define a language-related loss term,
\begin{align}
\nonumber
\mathcal{L}_{\text{language}}(\hat C_{\Psi})  = \operatorname{CE}\left(p_{i+1}(\hat C_{\Psi}) ,  \operatorname{PLM}(t_i, [\bar C_p, C_i])\right),
\end{align}
which is the cross-entropy loss of the optimized PLM, as defined in Eq.~\ref{eq:plm_prompt}, with the unmodified PLM distribution.

  In order to have the generated text describe the set of images using fluent language, we solve the following optimization problem: 
\begin{align}
\nonumber
     \min_{\hat C_{\Psi}} \mathcal{L}(\hat C_{\Psi}) = \min_{\hat C_{\Psi}}  \mathcal{L}_{\text{vision}} (\hat C_{\Psi}) + \lambda \mathcal{L}_{\text{language}} (\hat C_{\Psi})  
\end{align}
 where hyper-parameter $\lambda$ calibrates the trade-off between relevance to the video and language fluency.
The optimization process occurs during autoregression inference, generating sentences iteratively. We detail this process next. 

\paragraph{Evolving Pseudo-Tokens Optimization}

The optimization occurs during the generation of the entire sentence, increasing image correspondence by applying iterations. Notably, we do not use annotations, nor are any parameters fine-tuned in a separate phase.  For each generated token, we perform one optimization step 
 \begin{align}
    \hat C_{\Psi} \longleftarrow \hat C_{\Psi} + \alpha \frac{\nabla_{C_{\Psi}}\mathcal{L}(\hat C_{\Psi})}{\|\mathcal{L}(\hat C_{\Psi})\|^2},
\end{align}
and continue the generation process. We stop the process and start a new sentence upon reaching a dot token, keeping the optimized pseudo-tokens. This optimization process is autoregressive in its nature, and, therefore, the first words are generated with preliminary context. Though early sentences usually have good language, each successive sentence increases the ability to ground objects. 

As a result, the process shifts from a general discussion to a more concrete explanation, see Sec.~\ref{sec:exp}.  Throughout our experiments, we employ 16 generation iterations. Each time a new sentence is generated, we pick a prefix-token from the set $\mathcal{P}$ at random, which acts as data augmentation, see Sec.~\ref{sec:image_captioning} for an analysis of using a random prefix. 

\paragraph{CLIP-based Frame Sampling for Video Captioning} Any set of images, including video frames, can be described with our method. Adapting our model for video captioning begins by sampling three frames every second. In order to avoid repetition and to capture diverse frames, we further subsample the frames with a CLIP-based strategy. 

We begin with the first frame as an anchor. The next frame (a third of a second later, due to the initial sampling) is included in the set if its distance from the anchor, in the space defined by the CLIP image encoder, exceeds a certain threshold $\lambda_{\text{frame}}$. The newly selected frame becomes the new anchor and the selection process continues in the same manner. See appendix Fig.~\ref{fig:picked_frames} for qualitative results. We use $\lambda_{\text{frame}}$=$0.9$ in all our experiments.

\begin{figure*}[t]
	\centering
    \includegraphics[width=1\linewidth]{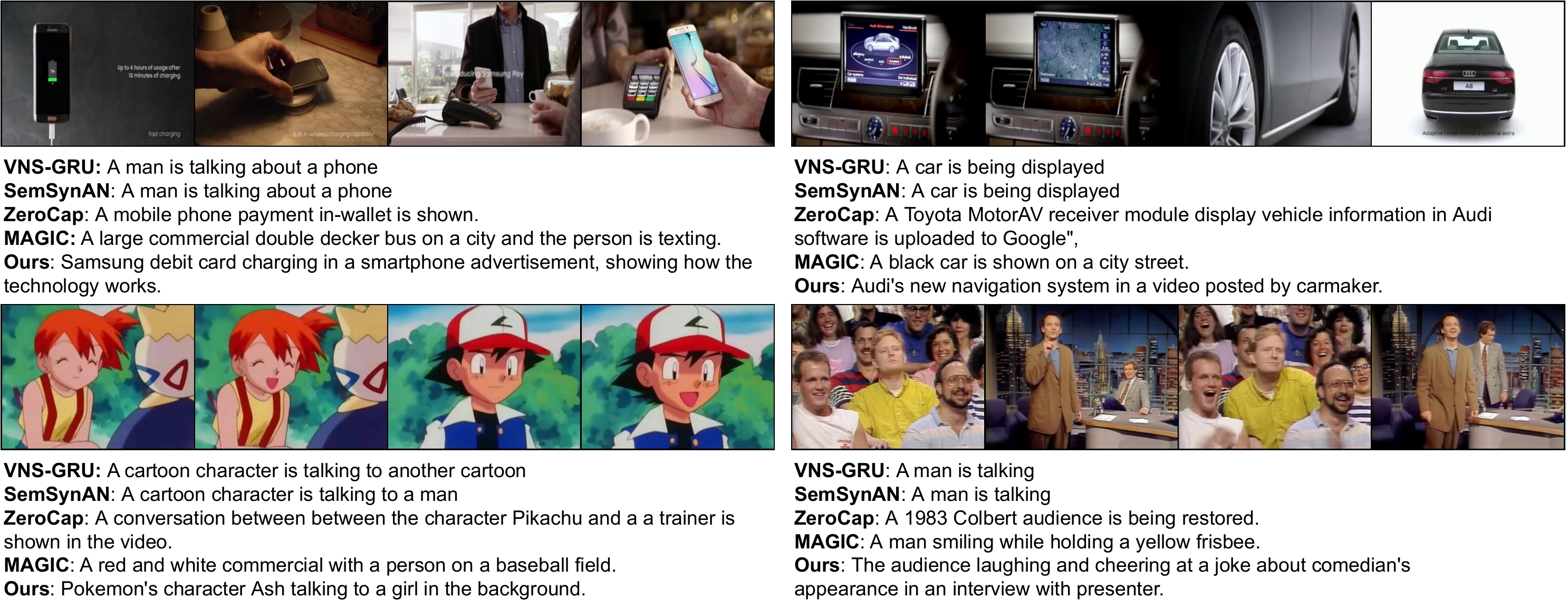}\vspace{-5pt}
    \caption{Examples of our video captions with two types of baselines: (i) the supervised methods SemSynAN and VNS-GRU; and (ii) the zero-shot methods ZeroCap and MAGIC. Notably, our method grounds objects from different frames and exhibits real-world knowledge. The 1st and 2nd rows provide examples of real-world knowledge.}\vspace{-5pt}
    \label{fig:qual_video}
    
\end{figure*}

\begin{figure*}[t]
	\centering
    \includegraphics[width=1\linewidth]{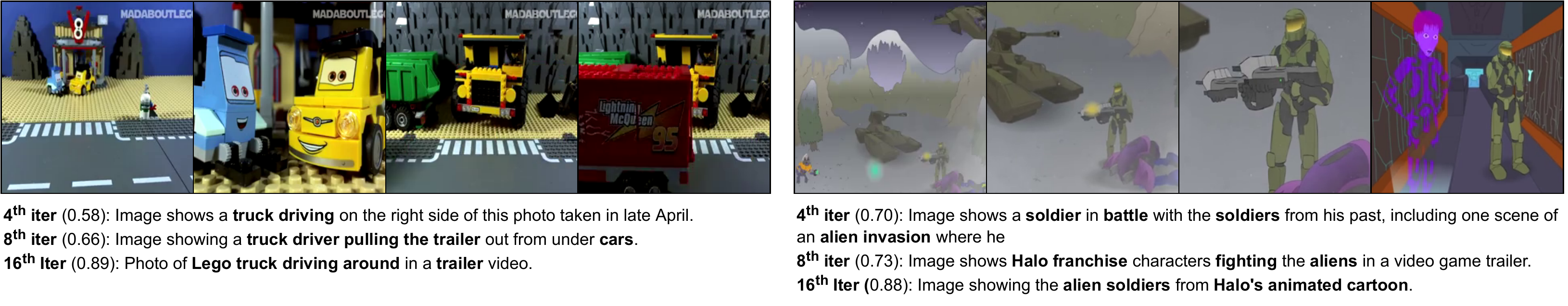}\vspace{-5pt}
    \caption{Evolution of video captions. We show the sentence with the highest CLIP score at different generation iterations. Grounded words are highlighted. }\vspace{-15pt}
    \label{fig:evolution_video}
\end{figure*}

\section{Results}
\label{sec:exp}
For all experiments, the following settings are used (see appendices for parameter sensitivity and ablations): We set $\lambda$ to $0.8$. During sentence generation, we pick one of the top-3 tokens at random. To avoid long repetitive sentences, the number of generated tokens per sentence was limited to 20. To avoid generating irrelevant entities, such as names, we reduce by 1 the logits of tokens with uppercase letters. Using a single Titan X GPU, each sentence takes 5 seconds to generate. All 16 sentence-generating iterations take approximately a minute and a half.

Two video datasets are used: MSR-VTT~\cite{xu2016msr-vtt} and MSVD~\cite{2016arXiv160906782W}. MSR-VTT is a large-scale dataset with about 50 hours of videos divided into 10,000 videos with 20 descriptions each. It includes a variety of categories, such as video games and TV shows. The test set consists of 2,990 videos. MSVD contains 1,970 short video clips, 670 of which are dedicated for testing.  All experiments are carried out on the test set.

We use two types of metrics: (i) Supervised metrics that measure text correspondence to human references: BLEU~\cite{Papineni2001BleuAM}, METEOR~\cite{Banerjee2005METEORAA}, CIDEr~\cite{Vedantam2014CIDErCI}, SPICE~\cite{anderson2016spice}.  Lastly, $\text{CLIP-S}^{\text{Ref}}$~\cite{hessel2021clipscore} measures semantic similarity by utilizing CLIP's textual encoder. (ii) Unsupervised metrics that are computed without referring to the human annotation. Relatedness to the visual cue is measured by averaging CLIP or BLIP image similarity scores to the generated sentence across the frames. Relatedness to the video is measured by  the VideoCLIP~\cite{xu2021videoclip} video-to-text distance metric (``Retrieval'' in the results table). Language quality is estimated using the perplexity score of the generated caption, employing BERT~\cite{devlin2018bert}.

\begin{table*}[t]
    \centering

	    \begin{tabular}{lcccccccc} 
	    \toprule
	    & \multicolumn{5}{c}{\textbf{Supervised Metrics}}  & \multicolumn{2}{c}{\textbf{Unsupervised Metrics}}  \\ 	
		\cmidrule(lr){2-6} 	\cmidrule(lr){7-8}    
		Method & B@4 & M & C & S & $\text{CLIP-S}^{\text{Ref}}$ & CLIP-S & PP \\ 
		\midrule
  		VinVL~\cite{zhang2021vinvl} & 0.41&	0.311 &	1.409&	0.252&	0.83&	0.780&	24.16 \\
  		\cmidrule(lr){2-8} 
    	&\multicolumn{6}{c}{\textit{Zero-Shot Methods}}\\
    	\cmidrule(lr){2-8} 
		ZeroCap~\cite{tewel2021zero} &	0.029&	0.12&	0.131&	0.055&	0.778&	0.870&	25.737 \\
		MAGIC~\cite{2022arXiv220502655S} &	\textbf{0.129}&	\textbf{0.174}&\textbf{	0.493}&	\textbf{0.113}&	0.763&	0.737&	37.126 \\
		\midrule
        Ours &	0.022&	0.127&	0.172&	0.073&	\textbf{0.798}&	\textbf{0.885}&   \textbf{19.049} \\
		\bottomrule
	    \end{tabular}
	\caption{Quantitative results for image captioning methods. We evaluate supervised metrics that measure text correspondence to human references and unsupervised metrics that are computed without referring to the human annotation.}
	\label{tab:COCO_SOTA}
\end{table*}

\begin{table}[t]
\centering

\begin{tabular}{lcc} 
	    \toprule  
		\textbf{Method} & \textbf{Image} & \textbf{Video} \\ 
		\midrule
		MAGIC~\cite{2022arXiv220502655S} & 1.65 & 1.77  \\
		ZeroCap~\cite{tewel2021zero} & 2.52 & 2.30  \\
		Ours & \textbf{4.01} & \textbf{4.14}  \\
		\bottomrule
\end{tabular}
\caption{Mean Opinion Score (MOS, scale of 1--5) for caption quality using real-world images and videos.}\vspace{-10pt}
\label{tab:user_study_table}
\end{table}

\paragraph{Quantitative analysis}

In Tab.~\ref{tab:video-cap}, we compare our approach with supervised state-of-the-art baselines for video captioning. We also compare it with zero-shot video captioning baselines we created by modifying CLIP-based zero-shot image captioning methods: ZeroCap~\cite{tewel2021zero}, a zero-shot method for image captioning, which also optimizes the generated sentence during inference. We adapt their method from image captioning to video captioning by replacing their single image CLIP loss with ours (i.e., a sum of CLIP losses for each frame). MAGIC~\cite{2022arXiv220502655S}, another zero-shot method for image captioning, which uses MAGIC scores, i.e., a CLIP-based measure of how closely a sentence ending with a given token matches an image, to skew the next-token distribution of a pre-trained language model to match a given image. To adapt their model to videos, we aggregate the CLIP score of all sampled frames to calculate the magic score before applying a softmax.

As expected, the supervised models VNS-GRU~\cite{chen2020delving} and SemSynAN~\cite{Perez-Martin_2021_WACV} perform significantly better on supervised metrics, based on correspondence to human references. However, when considering semantic relatedness to annotations (i.e., CLIPScoreRef), our method is better (0.785 vs. 0.739 and 0.733). We next look at unsupervised metrics. BLIP-Score suggests that our text is more relevant to the frames (0.675 vs. 0.623 and 0.608). Furthermore, when considering the entire video temporally, our method has the lowest VideoCLIP text-to-video distance (0.504 vs. 0.446 and 0.437). To understand the source of the weakness of the supervised methods, we measure the novelty of the generated sentences.  Aggregated over the entire MSR-VTT test set, our method has a vocabulary size of 4,372. In contrast,  SemSynAN and VNS-GRU use only 359 and 435 words, respectively, with roughly 40\% of the generated sentences existing in the training set.

Focusing next on zero-shot methods, Zero-Cap has higher CLIP-related scores than the supervised method. However, in all other metrics, it falls short. One of the main limitations of this technique is that the generated token is optimized until convergence in a greedy manner, without the option of altering past tokens. In contrast, our approach optimizes the generation of an entire sentence, which is beneficial for coherent generation from noisy signals resulting from multiple visual cues. 
\begin{figure*}[t]
	\centering
    \includegraphics[width=1\linewidth]{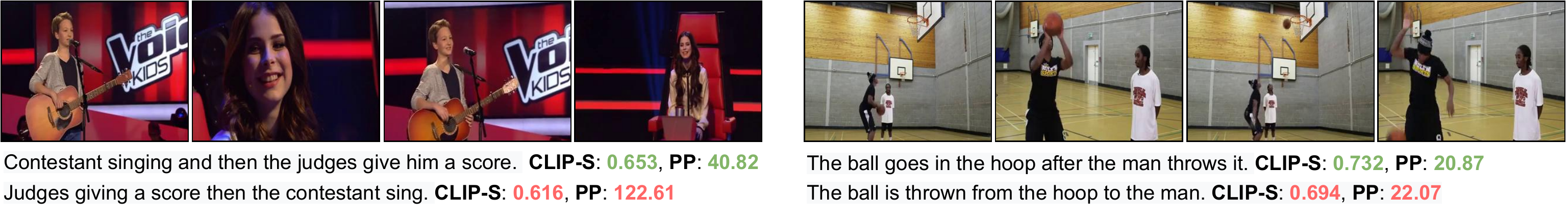}\vspace{-5pt}
    \caption{Counterfactual examples, which reorder the events in generated sentences. The analysis shows that temporal knowledge is embedded in the language and in visual cues.}\vspace{-5pt}
 
    \label{fig:temporality_ablation}
    
\end{figure*}
MAGIC samples a next token at each generation step from a distribution that combines both CLIP potentials and the distribution computed using a language model. To ensure that the resulting captions lie in the text domain of image captions, they fine-tune the language model on the text corpus of MS-COCO captions. While MAGIC is comparable to our method with regards to the supervised metrics, it falls short in all other metrics.
We hypothesize that MAGIC's per token greedy decoding harms the ability to handle the noisy signal of multiple visual cues found in a video.

\paragraph{Qualitative Analysis:} In Fig.~\ref{fig:qual_video}, we show examples of our video captions. First, we evaluate captions created using supervised methods. Despite having significantly different architectures, the supervised methods produce very similar captions. Although supervised methods may be on topic, their grounding capabilities are limited to relatively abstract objects. For example, the supervised method recognizes that the video is about a car or phone in the first row. However, it misses the true intent of the video.  

Our model displays real-world knowledge. For instance, it can detect brands. The video in the first row, on the left, shows Samsung's advertisement (i.e., Samsung Galaxy) about credit card payments, which is recognized by our method. The video on the right shows an Audi navigation commercial. Our model identifies the Audi brand and the video's purpose. The video in the 2nd row, on the left, demonstrates TV animation. Our method recognizes the character Ash from Pokemon. Next, we demonstrate that our model can recognize events from various frames. In the 2nd row, right side, our model describes the entire scene, both the audience and the comedian being interviewed.

We also compare our method to two other zero-shot methods, ZeroCap and MAGIC. We find that captions generated by other methods are often broad, failing in videos that require real-world knowledge or aggregating information from multiple frames. For instance, both methods miss the Samsung brand. Furthermore, they often identifies a related but wrong entity (e.g., Pickachu in the Pokemon video). Often, they miss events from different frames, e.g., they do not describe the comedian or the interviewer in the right video on the 2nd row.

In Fig.~\ref{fig:evolution_video} we evaluate how sentences evolve during the inference process. We first note that, quantitatively, the CLIP score increases between generation iterations, e.g., on the left, at the fourth sentence iteration, the clip score is 0.58, while at the 16th iteration, the score is 0.89. This improvement can also be seen qualitatively. In the left video, we see that the sentence grounds the truck, which is visible in all frames, after four iterations. The model grounded both the truck and trailer in its eighth iteration. Only after the 16th iteration does the model recognize it as a Lego truck. We note an interesting failure case in which, after 16 iterations, the model incorrectly identifies the type of video as a trailer. The video on the right shows similar behavior. In the 16th sentence iteration, the CLIP score increased from 0.70 to 0.88. In the fourth iteration, the video was grounded to more abstract objects (e.g., soldier, battle, alien), while the eighth iteration identified the characters from Halo. As a final step, the model figures out that the video is an animated cartoon in the 16th iteration. 

Tab.~\ref{tab:COCO_SOTA} compares our method to state-of-the-art image captioning approaches, ZeroCap, and MAGIC. These approaches optimize at the token level resulting in a drop in the PP metric. Our sentence-level optimizations generate more fluent captions compared to token-level optimizations. We also assess supervised metrics aimed at language correspondence against human references. MAGIC excels on the supervised metrics. As the method fine-tunes the PLM on the human references, it allows the language to relate to their style. When evaluating captions using CLIP-based scores, performance drops. Our next step is to explore the reason by using a small sample of images outside the COCO dataset. A significant advantage of zero-shot captioning methods is that they can describe images not included in the training set.  In Tab.~\ref{tab:user_study_table}, we assess the caption quality of zero-shot methods when presented with images or videos requiring real-world knowledge. We asked 20 annotators to consider three properties: language fluency, human-likeness, and grounding, and rank each caption between 1 (lowest score) and 5 (highest score). The test included 10 images randomly sampled from the web, and 10 videos randomly sampled from the MSR-VTT~\cite{xu2016msr-vtt} test set. Our approach is significantly better for videos (4.14 vs. 2.30)  and images (4.01 vs. 2.52). In  Fig.~\ref{fig:real_world_image}, we show different samples from the human evaluation. Generally, our method's captions provide a coherent description of the given entity. MAGIC, on the other hand, does not perform well. The proposed method may be hindered by the fine-tuning of the language model, which still interferes with the method's ability to discuss real-world entities. 

Furthermore, we stress-test our model by captioning sets of random images, which explores its ability to create a coherent text that describes unrelated images, see Sec.~\ref{sec:image_set_captioning} for further discussion.

\section{Discussion and Limitations}
\label{sec:discussion}

Despite showing video as the main application in most of our experiments, our method is invariant to the order of frames in the sequences. This is a result of using a zero-shot strategy, in which the underlying models are trained per frame. Examining the results, the generated captions display a logical order. This is not surprising, since ordering frames is typically not an extremely challenging visual understanding task, and is also used as a self-supervised task~\cite{xu2019self}. Evidently, the information that exists in both the Language Model and the CLIP network is sufficient for generating sentences that adhere to the natural order of events.

To further examine this, we created counter-factual sentences in which the order of events is altered, and measured the PLM and CLIP scores. As can be seen in the examples of Fig.~\ref{fig:temporality_ablation}, changing the order leads to a drop in both scores. For example, judges giving a score before a contestant sings has a higher perplexity, or a player's pose can be used to identify them as someone who throws a ball and not someone who catches it.

A limitation of large-scale models is that they can sometimes generate sexist, or otherwise toxic language. Before viewing the examples (see appendix Fig.~\ref{fig:toxicity}), readers are advised that they contain harsh language. CLIP and GPT-2 may be at fault because they use web-based, uncurated data~\cite{gehman2020realtoxicityprompts}. It is advisable to be aware of these weaknesses before deploying our method or any other method that uses these models.

\section{Conclusions}

We present a method for creating natural-sounding captions from a video. The method is a zero-shot method based on two frozen networks: a language model and an image-text matching model. The method is based on learning, for each input, a sequence of vectors that serve as pseudo-tokens that drive the generation process. An autoregressive process generates the caption while updating the pseudo-tokens. Once the caption is generated, the process repeats, using the learned pseudo-tokens as the starting point, leading to increasingly concrete and well-grounded captions. Our experiments show that our model generates novel captions that ground objects from multiple images into one coherent narrative.

\section*{Acknowledgments}
This project has received funding from the European Research Council (ERC) under the European Unions Horizon 2020 research, innovation programme (grant ERC CoG 725974). 

\bibliographystyle{abbrv}
\bibliography{main}

\appendix

\section{Appendix}
\label{sec:appendix}

In this appendix we provide additional results and ablations studies.

\begin{table*}[t]
    \centering
	\resizebox{\linewidth}{!}
	{
	    \begin{tabular}{lcccccccccc} 
	    \toprule
	    & & \multicolumn{6}{c}{\textbf{Supervised Metrics}}  & \multicolumn{3}{c}{\textbf{Unsupervised Metrics}}  \\ 	
		\cmidrule(lr){3-8} 	\cmidrule(lr){9-11}    
		Method & Prefix & B@4 & M & C & R & S & $\text{CLIP-S}^{\text{Ref}}$ & CLIP-S& BLIP-S & PP \\ 
		\midrule
		ZeroCap~\cite{tewel2021zero} & None	& 0.021	& 0.1 & 0.139 &	0.207 &	0.051	& 0.760	& 0.821	& 0.604& 109.959  \\
		ZeroCap~\cite{tewel2021zero} & `A'&	0.026&	0.116&	0.145&	\bf0.276&	0.054	&0.771&	0.845&	0.611&	33.661 \\
		ZeroCap~\cite{tewel2021zero} & `Image of a'&\bf	0.029&	0.12&	0.131&	0.268&	0.055&	0.778&	0.870&	0.605&	25.737 \\
		\midrule
        Ours & None&	0.024&	\bf0.127&	\bf0.200&	0.239&	0.071&	0.791&	0.852&	\bf 0.652&	20.412 \\
        Ours & Random&	0.022&	\bf0.127&	0.172&	0.228&	\bf0.073&	\bf0.798&	\bf0.885&	0.651&	\bf19.049 \\
		\bottomrule
	    \end{tabular}
    }
    \medskip
	\caption{Quantitative results for image captioning on the MS-COCO test set using different prefixes.}
	\label{tab:COCO}
\end{table*}

\section{Quantitative Results}

\subsection{Image Captioning}
\label{sec:image_captioning}
Tab.~\ref{tab:COCO} shows results for single-image captioning, evaluated on the MS-COCO test-set~\cite{ty2014coco}. We compare our method with another zero-shot method, ZeroCap. Unlike the baseline, our method is trained with a perturbed prefix instead of `Image of a'. As a result, our method is more robust to prefix changes. Notably, the perplexity score is significantly higher when the prefix is removed from the baseline sentence (109.959 vs. 25.737). Furthermore, compared to the baseline's 0.870 CLIP score, our method has a higher CLIP score of 0.885. CLIPScoreRef is also improved (0.778 vs. 0.798), which means that our caption matches human references better. In particular, we optimize complete sentences, resulting in a significant improvement in language fluency (19.049 vs. 25.737). 

\subsection{Image Set Captioning}
\label{sec:image_set_captioning}

In order to stress-test our method, we consider the task of captioning random image sets. The goal is to describe a set of images with one coherent sentence. We use the MS-COCO test-set~\cite{ty2014coco} for images. The number of images varies between one and four, one being the conventional image captioning task.

\begin{figure}[t]
\centering
\begin{subfigure}{0.32\textwidth}
    \includegraphics[width=\textwidth]{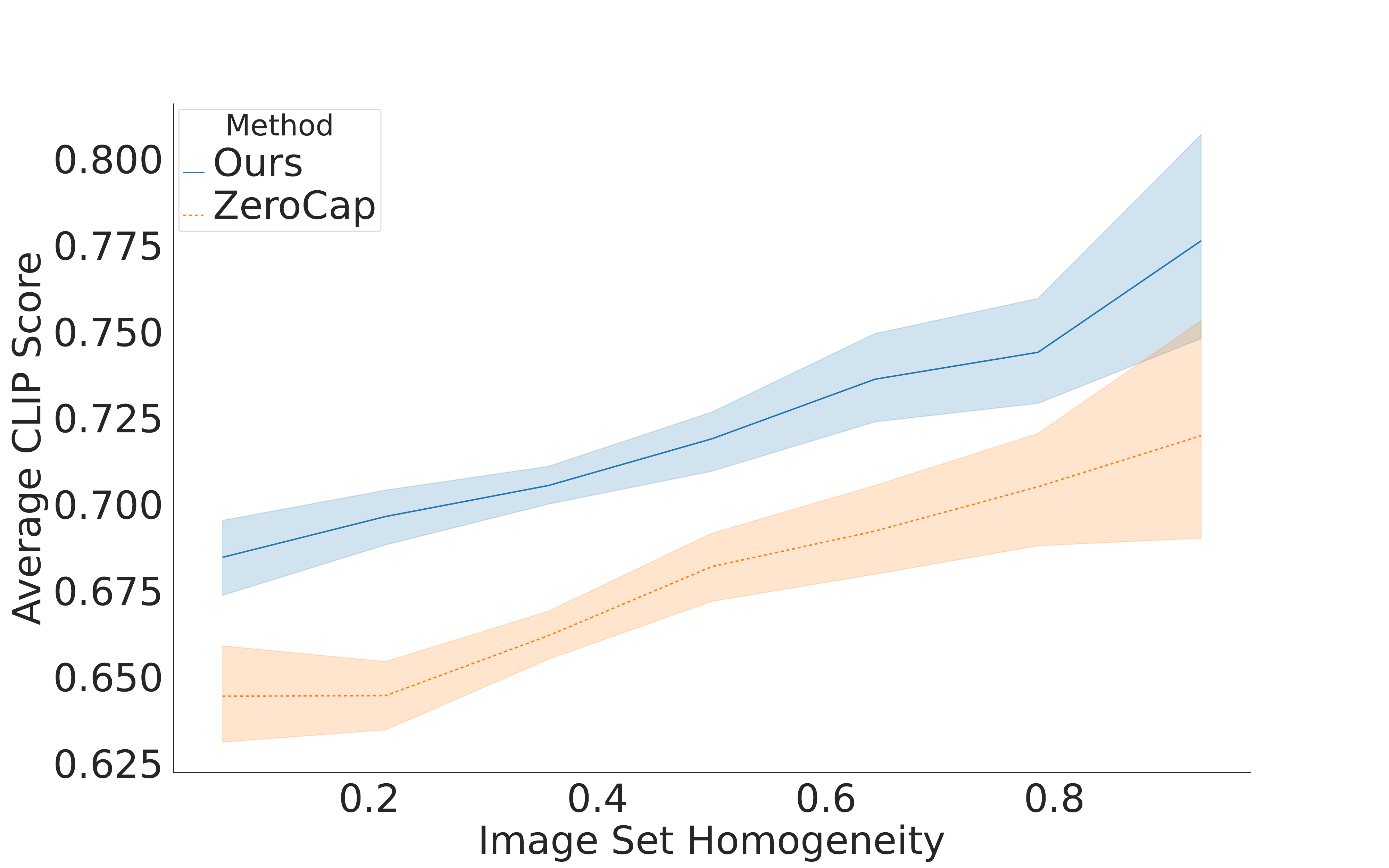}
    \caption{Two images}
\end{subfigure}
\hfill
\begin{subfigure}{0.32\textwidth}
    \includegraphics[width=\textwidth]{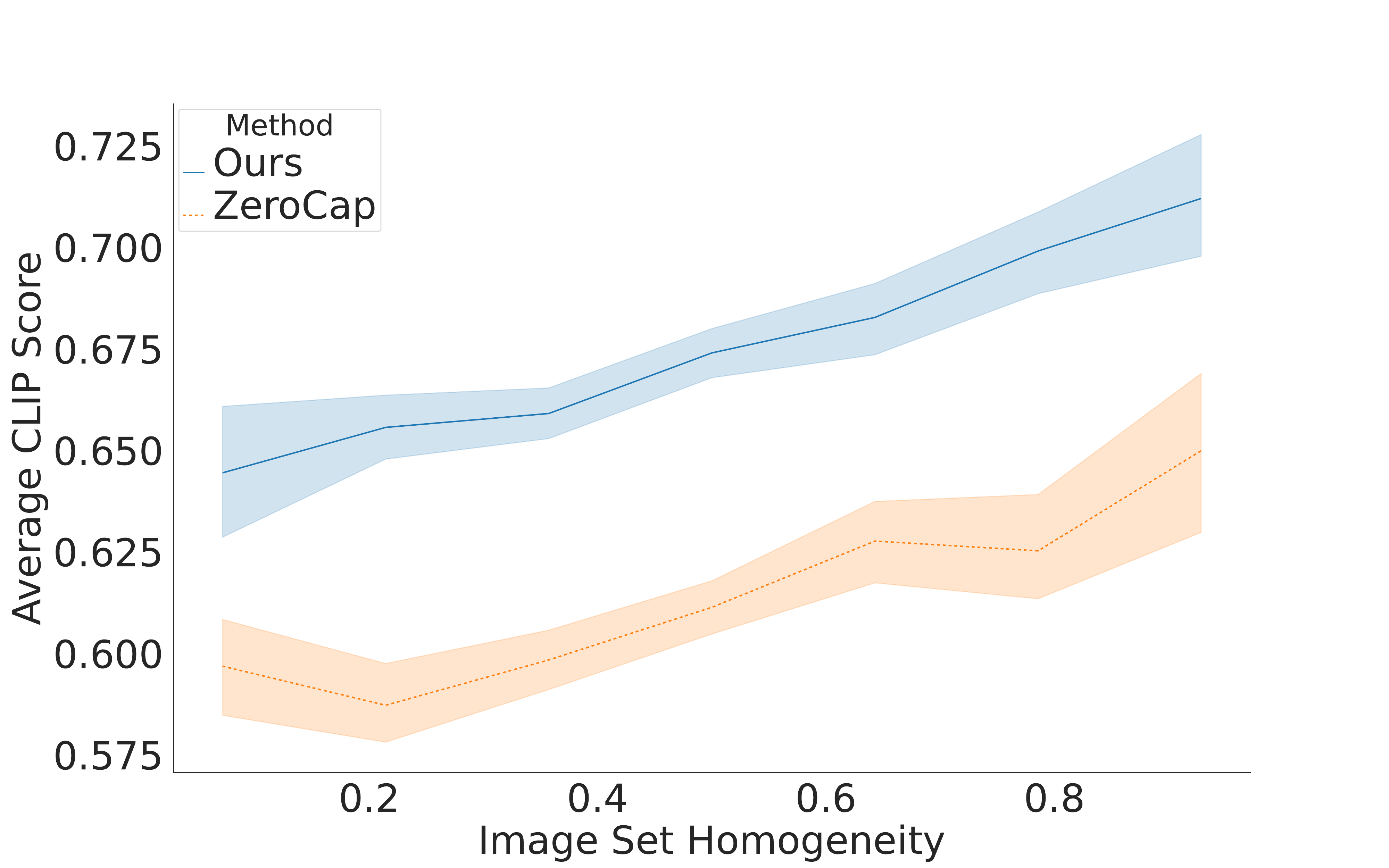}
    \caption{Three images}
\end{subfigure}
\hfill
\begin{subfigure}{0.32\textwidth}
    \includegraphics[width=\textwidth]{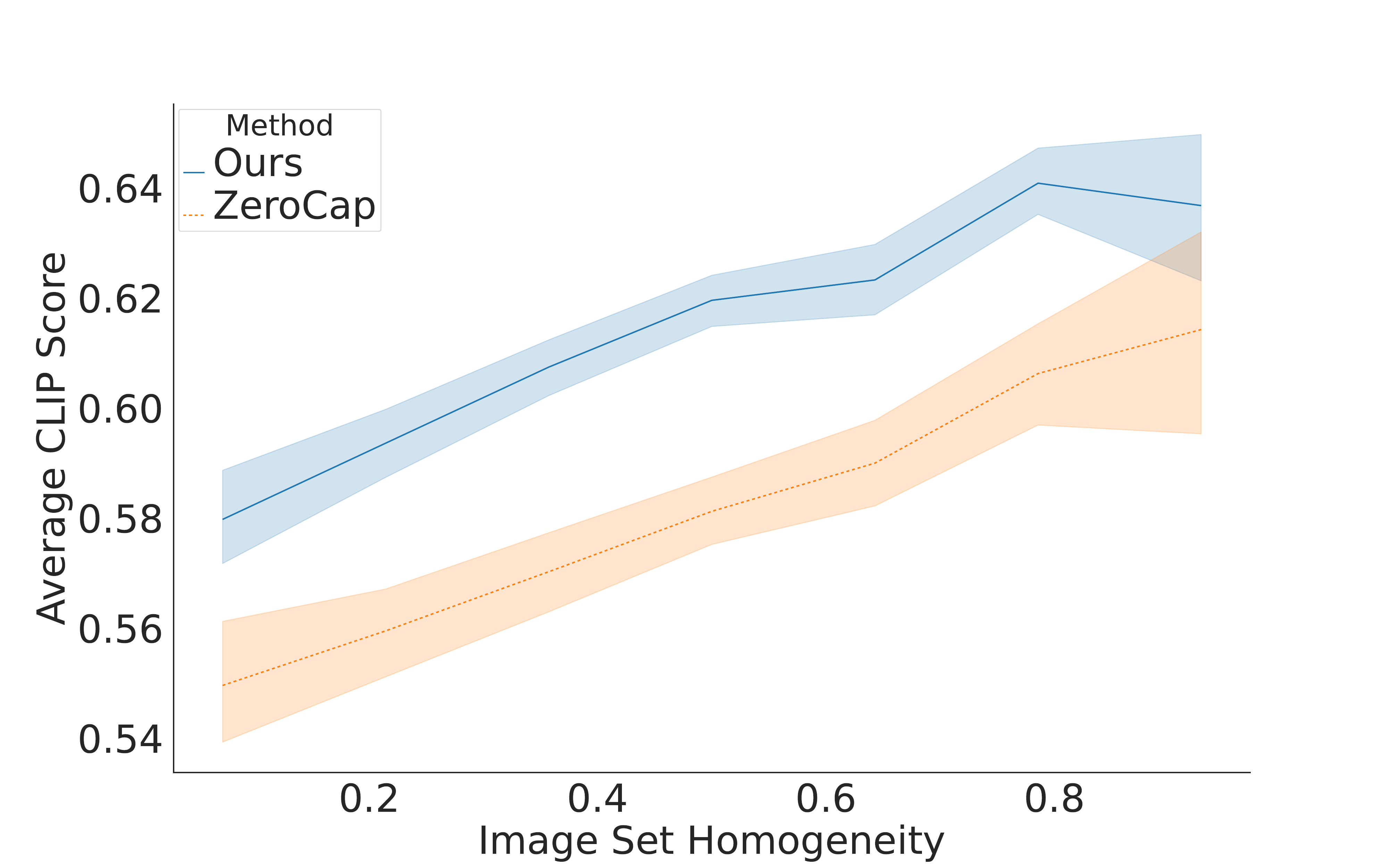}
    \caption{Four images}
\end{subfigure}
        
\caption{CLIP-score for different sets of images varying by size and set homogeneity.}
\label{fig:clip_score_homogeneity}
\end{figure}

It is expected that the more heterogeneous the image set, the harder it is to generate a coherent caption. To quantify this, we measure the homogeneity of a set by the average CLIP score between the human captions of each image and the rest of the images in the set. Intuitively, this tells us whether the images depict similar concepts or not. In addition to comparing sets by their size, we also differentiate them based on their level of homogeneity. The results are shown in Fig.~\ref{fig:clip_score_homogeneity}. The graphs reveal that the CLIP score increases with homogeneity, which is expected, since a single sentence can describe homogeneous images better.
Our approach has a higher CLIP score at all levels of homogeneity and for all set sizes.

\begin{figure}[t]
\centering
\begin{subfigure}{0.32\textwidth}
    \includegraphics[width=\textwidth]{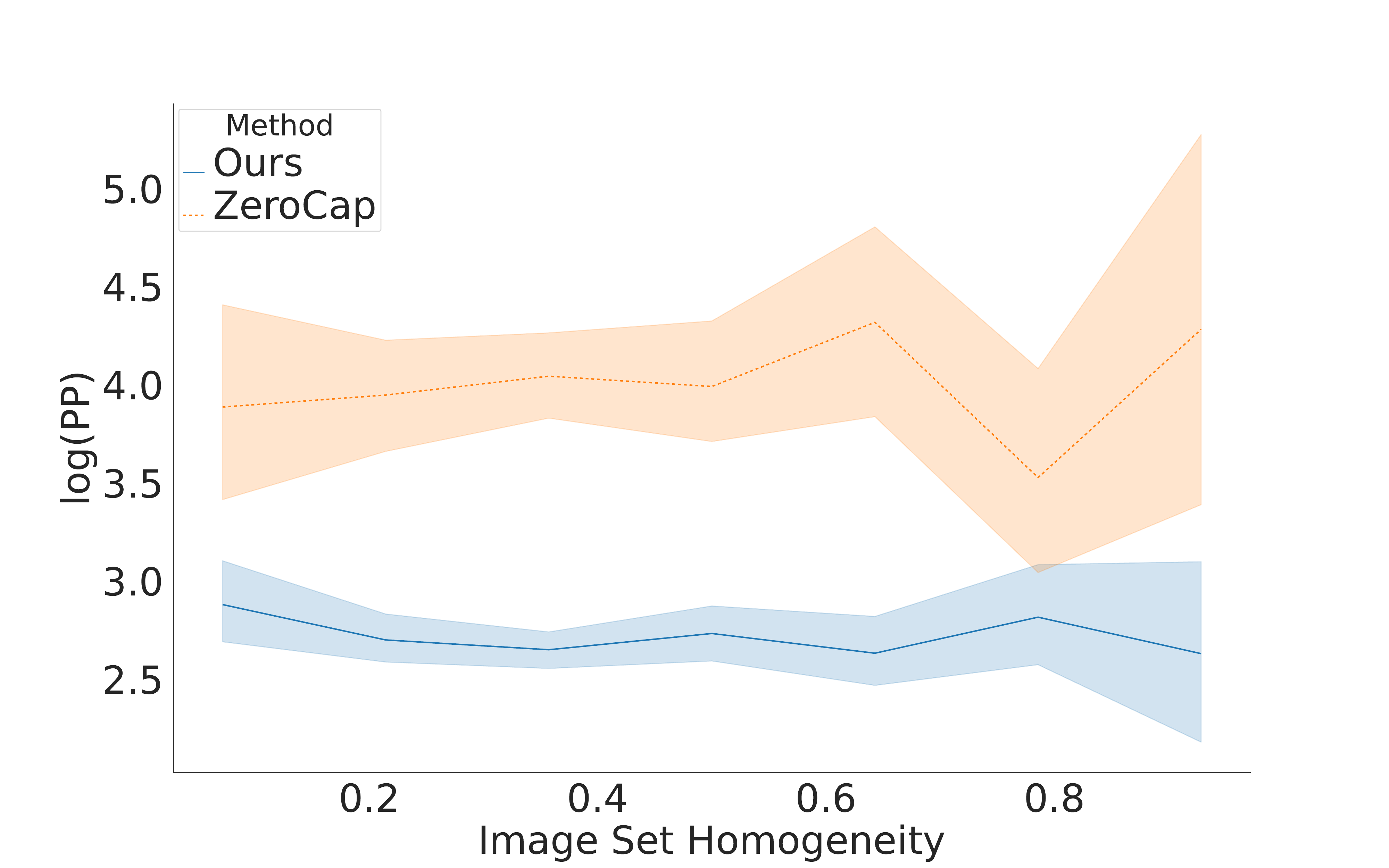}
    \caption{Two images}
\end{subfigure}
\hfill
\begin{subfigure}{0.32\textwidth}
    \includegraphics[width=\textwidth]{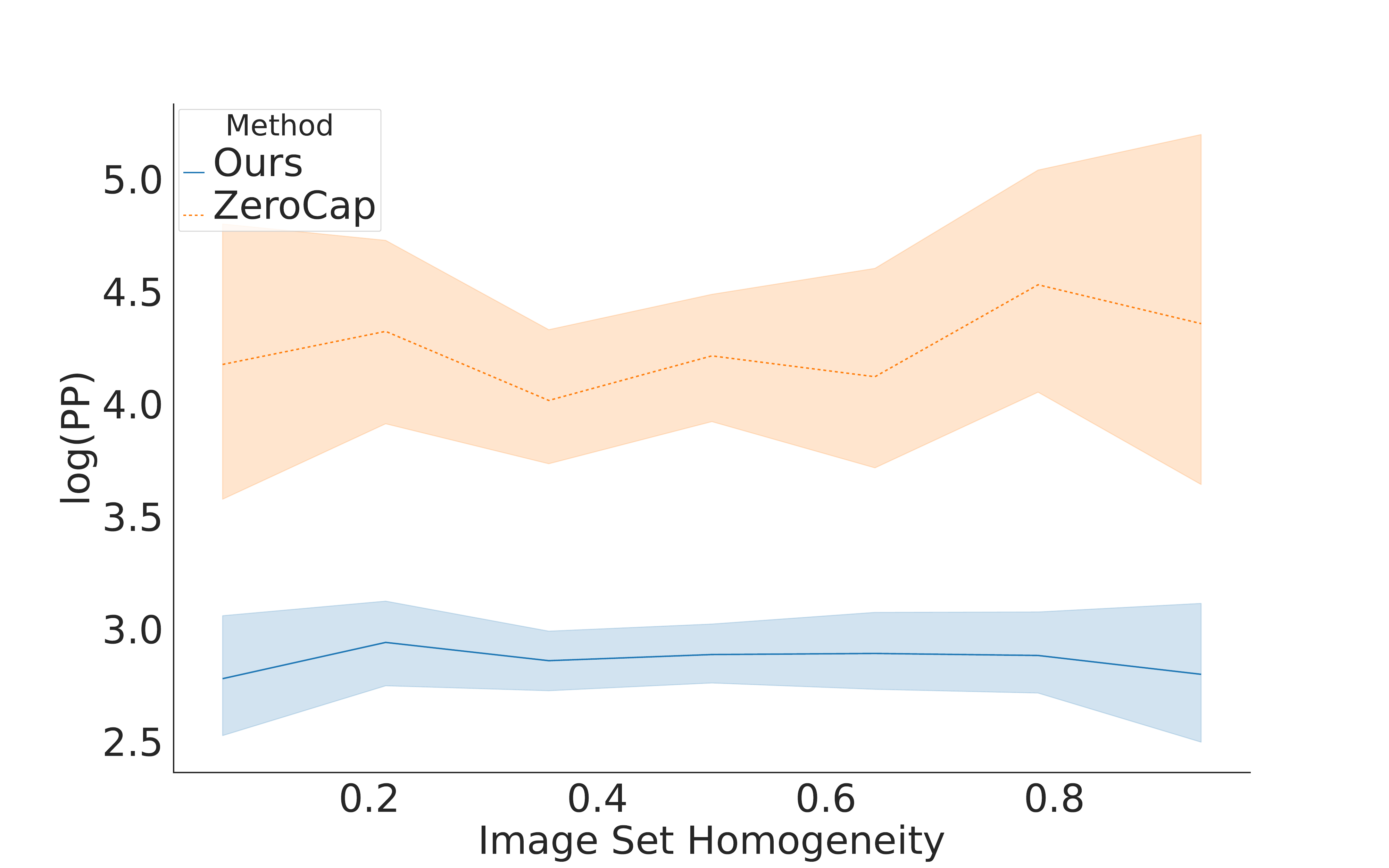}
    \caption{Three images}
\end{subfigure}
\hfill
\begin{subfigure}{0.32\textwidth}
    \includegraphics[width=\textwidth]{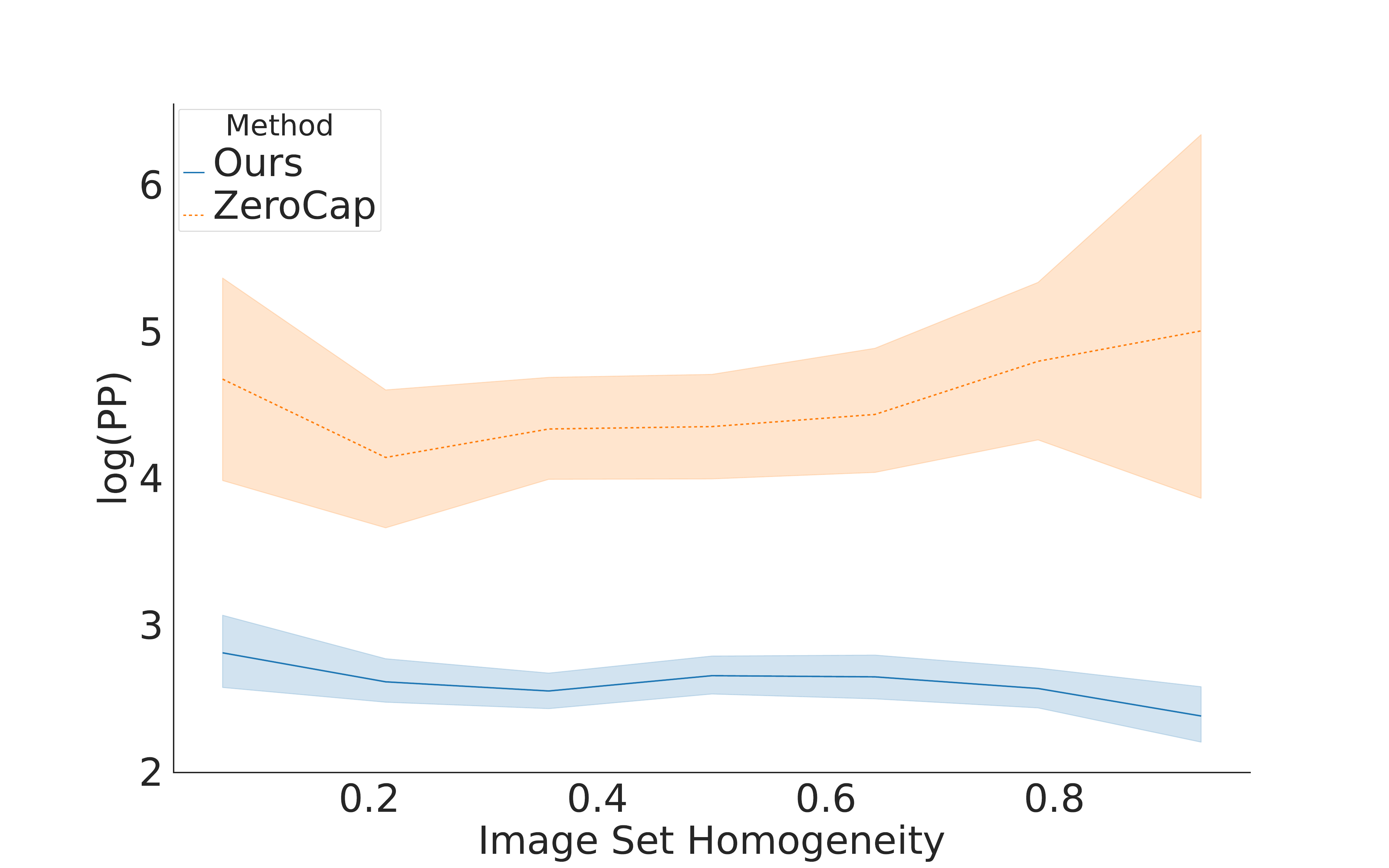}
    \caption{Four images}
\end{subfigure}
        
\caption{BERT-based perplexity score for different sets of images varying by size and set homogeneity.}
\label{fig:pp_score_homogeneity}
\end{figure}

In Fig.~\ref{fig:pp_score_homogeneity}, an experiment similar to the one above, based on BERT-based perplexity, is conducted to measure language quality. We find that our method produces much better sentences. A particularly interesting case is that of homogeneous pairs (i.e., homogeneity level of 0.8). Only in this case, which involves two very similar images, does ZeroCap perform as well as we do (logPP of 3.0 vs. 3.5). This highlights the ability of our approach to generate coherent sentences that describe a set of images across various challenges. 

\section{Ablation Study}

To assess different hyperparameters, we use the MSVD~\cite{2016arXiv160906782W} validation set, which consists of 100 videos. We examine two properties: (i) Video correspondence, which we examine with the Retrieval score, and (ii) language fluency, which we analyze with the BERT perplexity score. Additionally, we report CLIP Score and BLIP score, which measure image correspondence with the selected frames. 

\begin{figure}[p]
\centering
\begin{subfigure}{0.47\textwidth}
    \includegraphics[width=\textwidth]{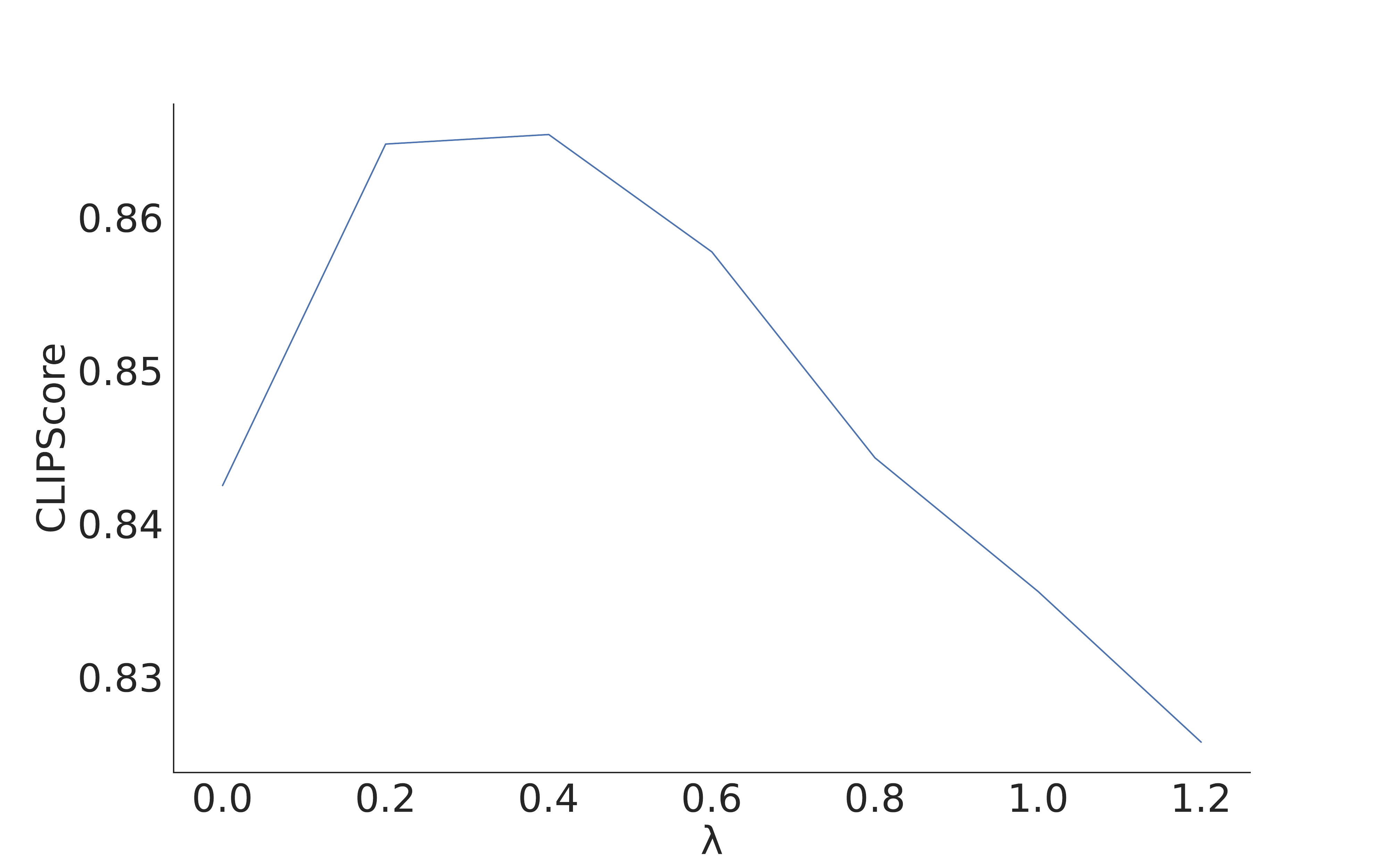}
\end{subfigure}
\hfill
\begin{subfigure}{0.47\textwidth}
    \includegraphics[width=\textwidth]{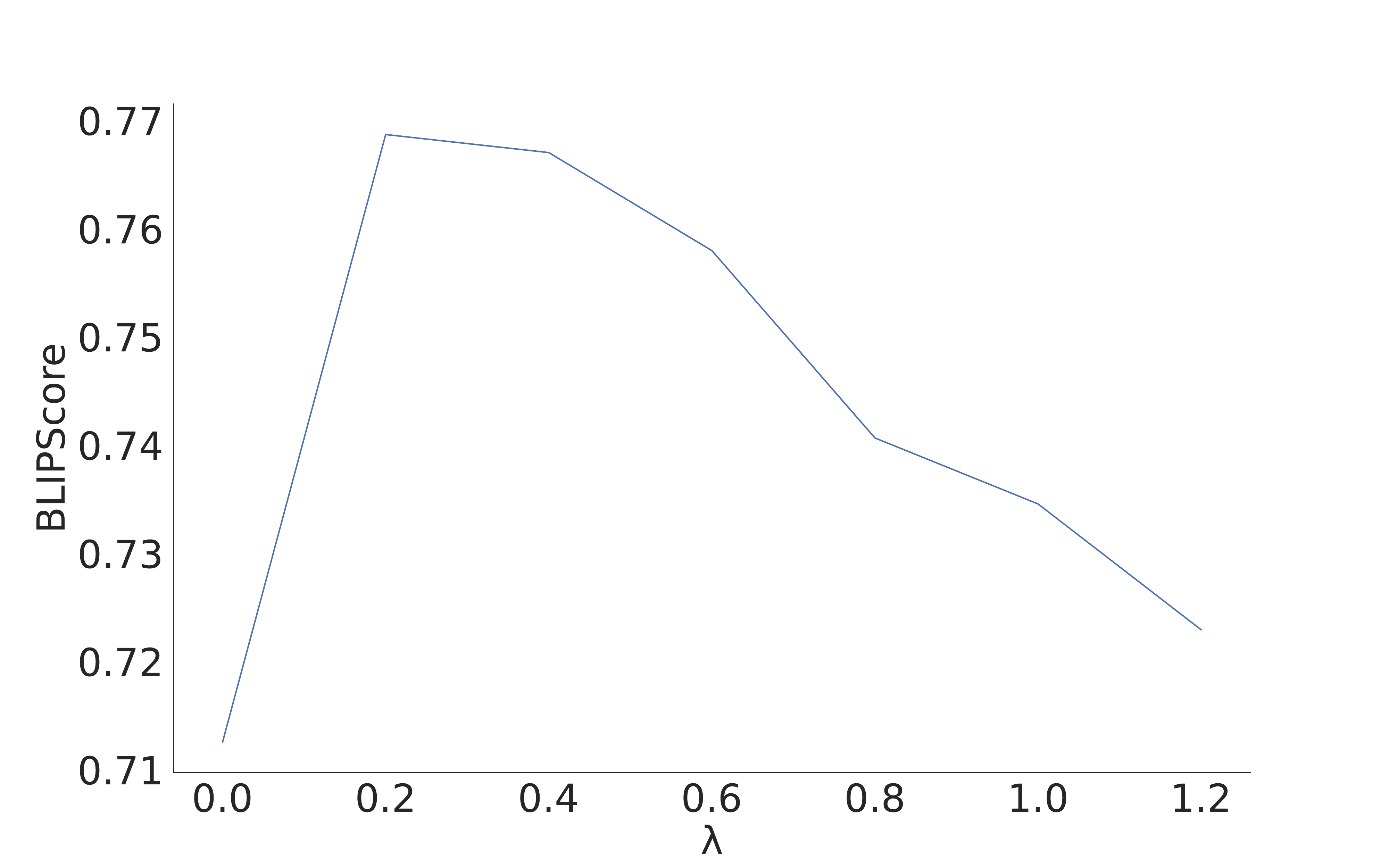}
\end{subfigure}
\hfill
\begin{subfigure}{0.47\textwidth}
    \includegraphics[width=\textwidth]{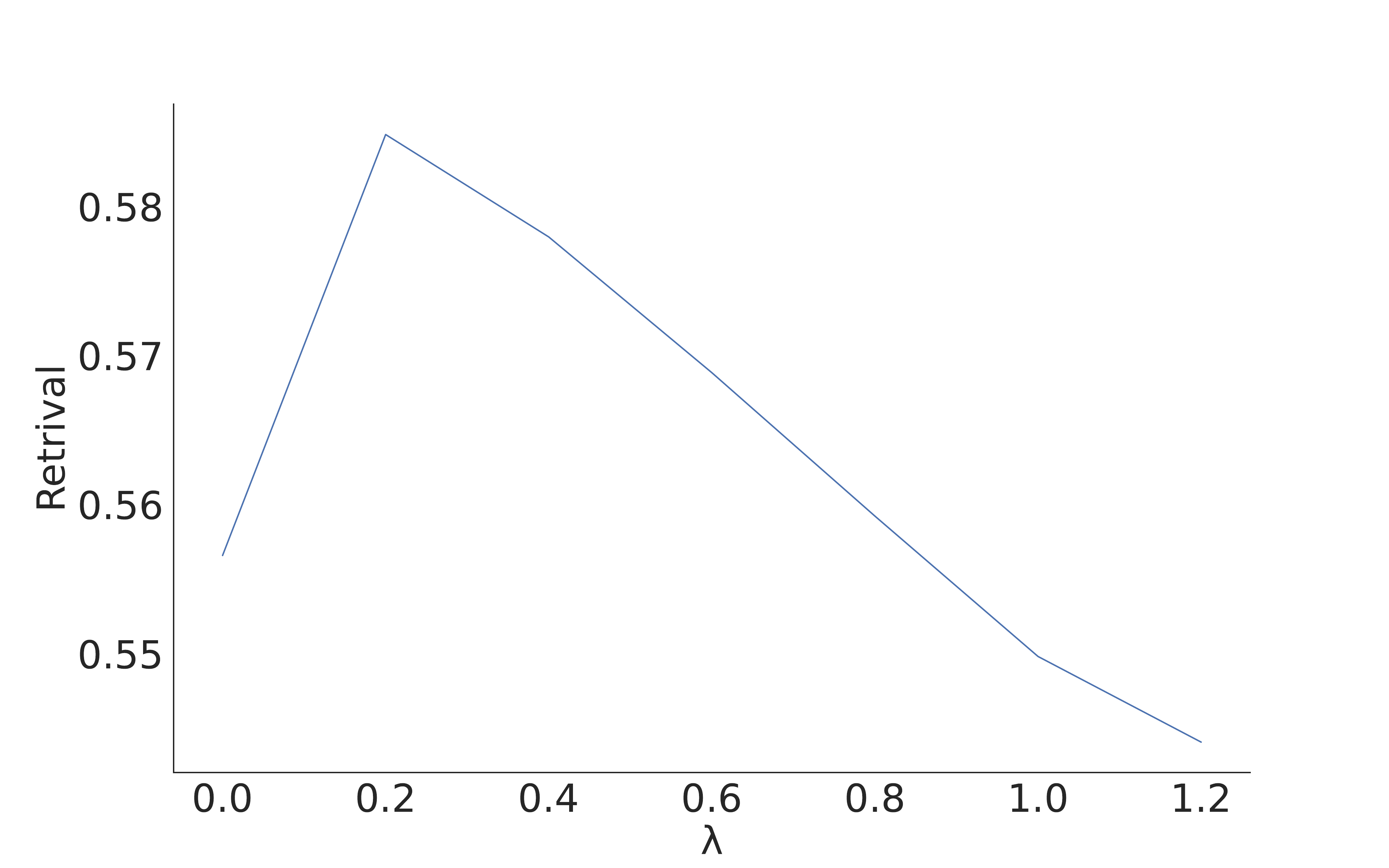}
\end{subfigure}
\hfill
\begin{subfigure}{0.47\textwidth}
    \includegraphics[width=\textwidth]{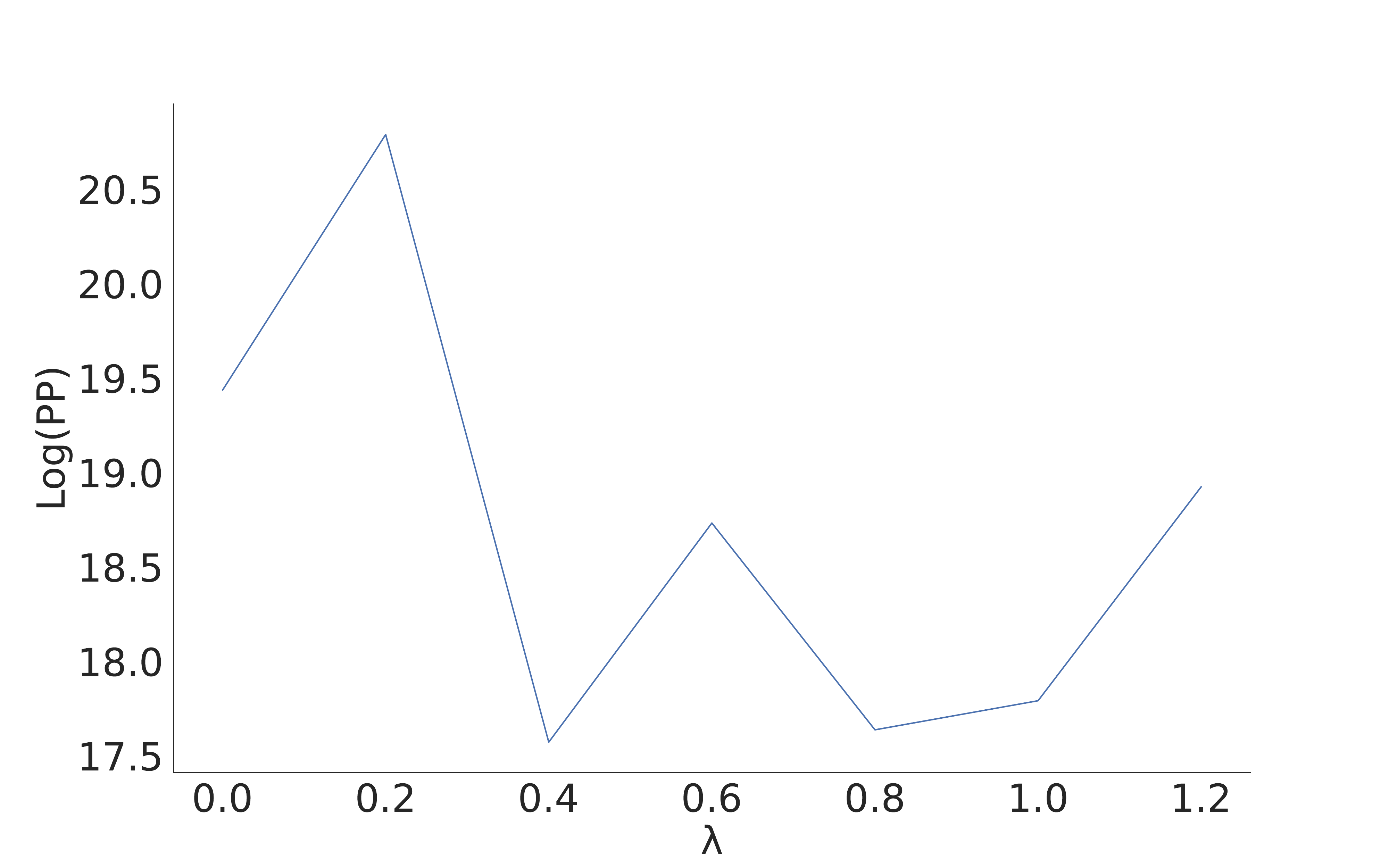}
\end{subfigure}
\hfill
\caption{Ablation study for the hyper-parameter $\lambda$.}
\label{fig:lambda}
\end{figure}

In Fig.~\ref{fig:lambda}, we study different values for $\lambda$, which controls the trade-off between CLIP loss ($\mathcal{L}_{\text{CLIP}}$) and language fluency loss ($\mathcal{L}_{\text{PLM}}$). Increasing the value of $\lambda$ decreases the Retrieval score. Our results show that  $\lambda=0.8$  provides a good trade-off between image correspondence and language fluency (i.e., low perplexity).

\begin{figure}[t]
\centering
\begin{subfigure}{0.47\textwidth}
    \includegraphics[width=\textwidth]{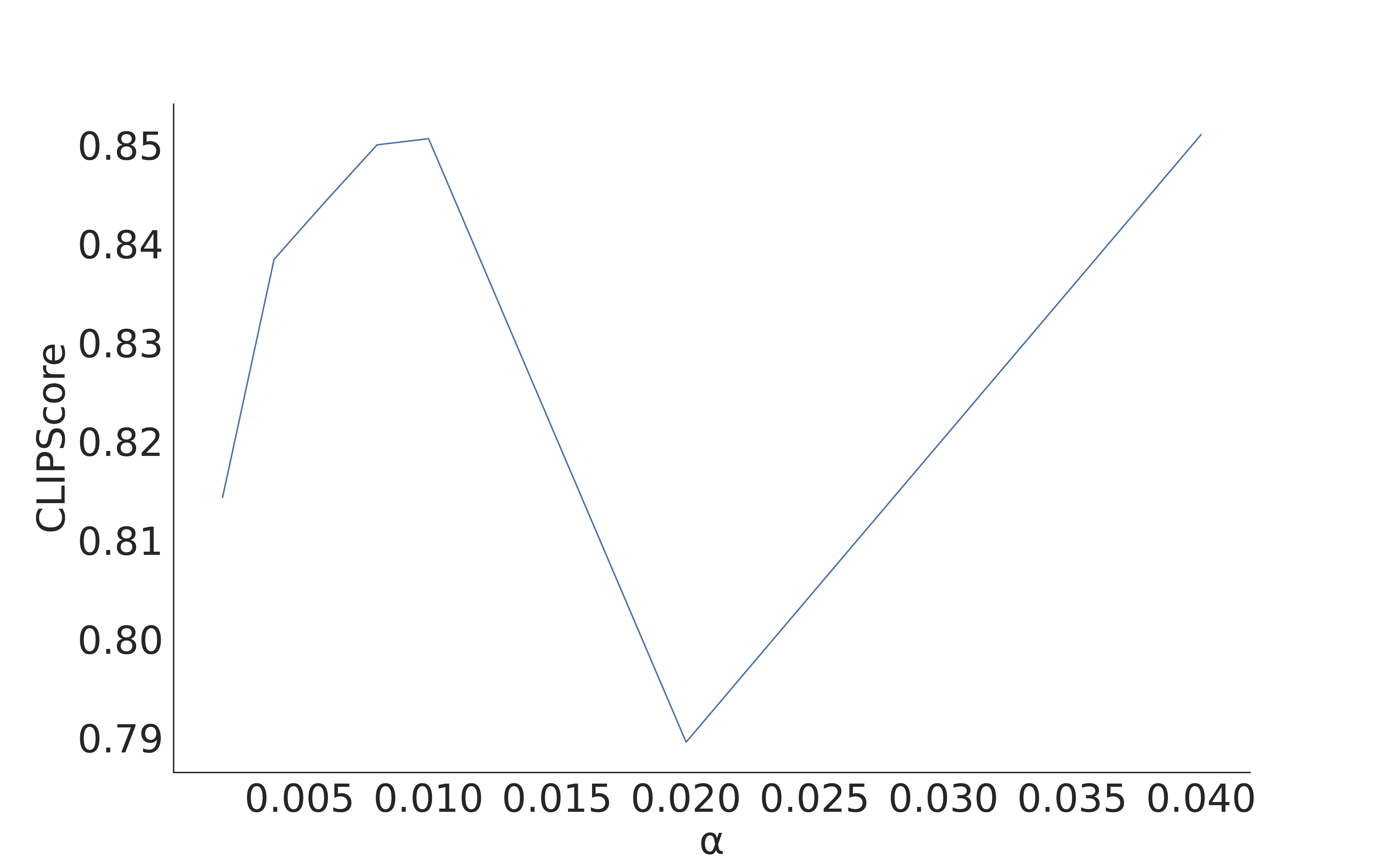}
\end{subfigure}
\hfill
\begin{subfigure}{0.47\textwidth}
    \includegraphics[width=\textwidth]{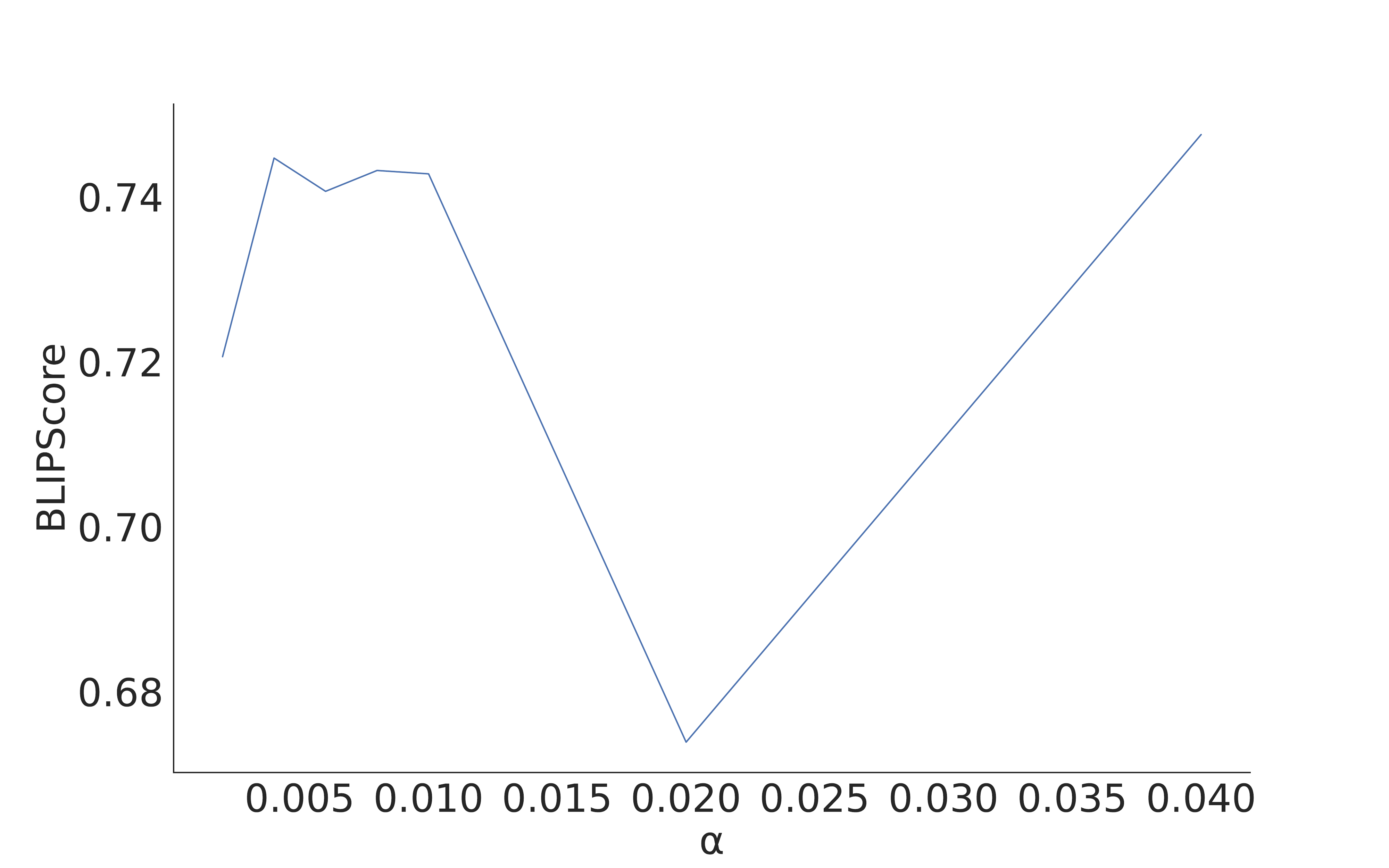}
\end{subfigure}
\hfill
\begin{subfigure}{0.47\textwidth}
    \includegraphics[width=\textwidth]{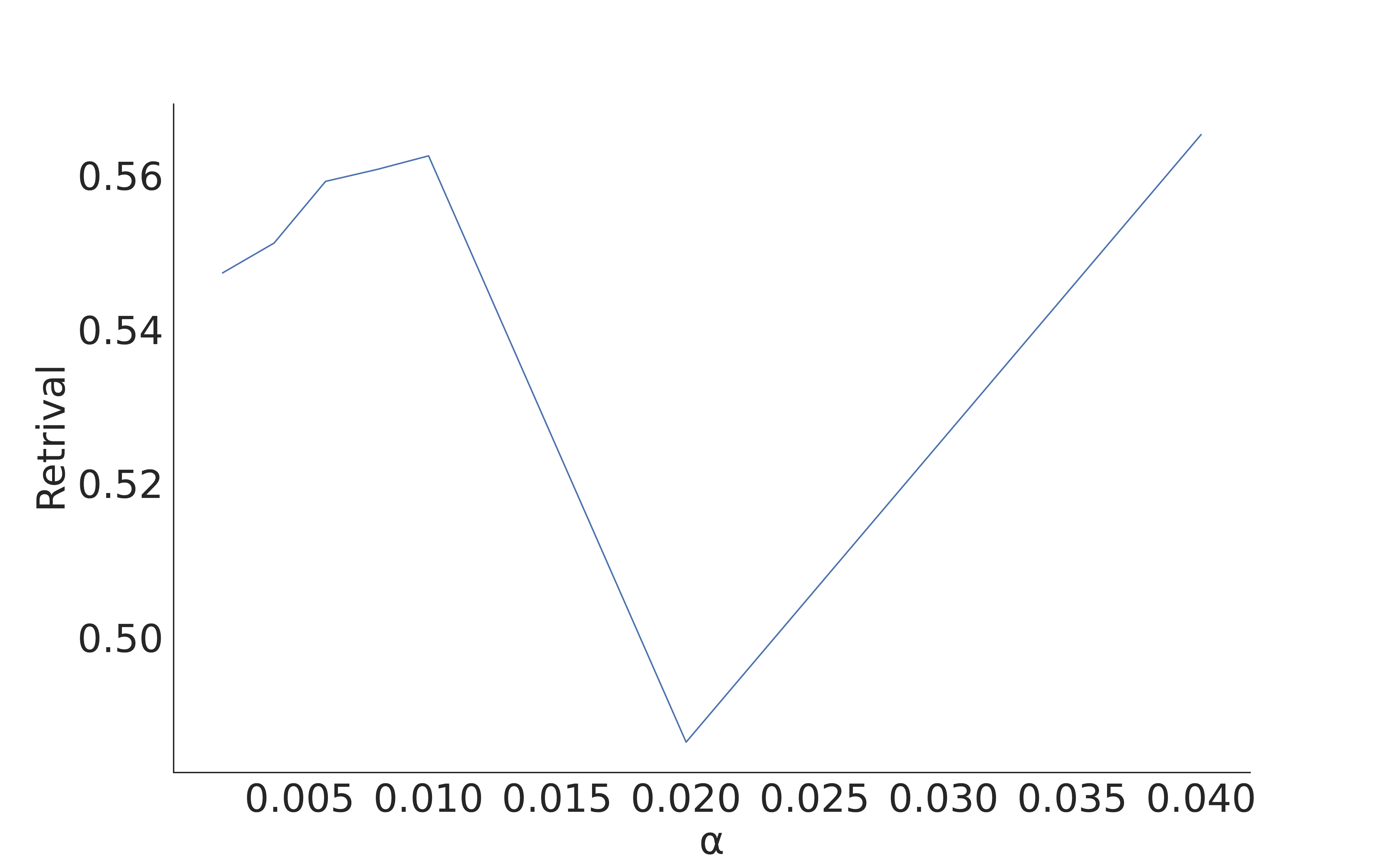}
\end{subfigure}
\hfill
\begin{subfigure}{0.47\textwidth}
    \includegraphics[width=\textwidth]{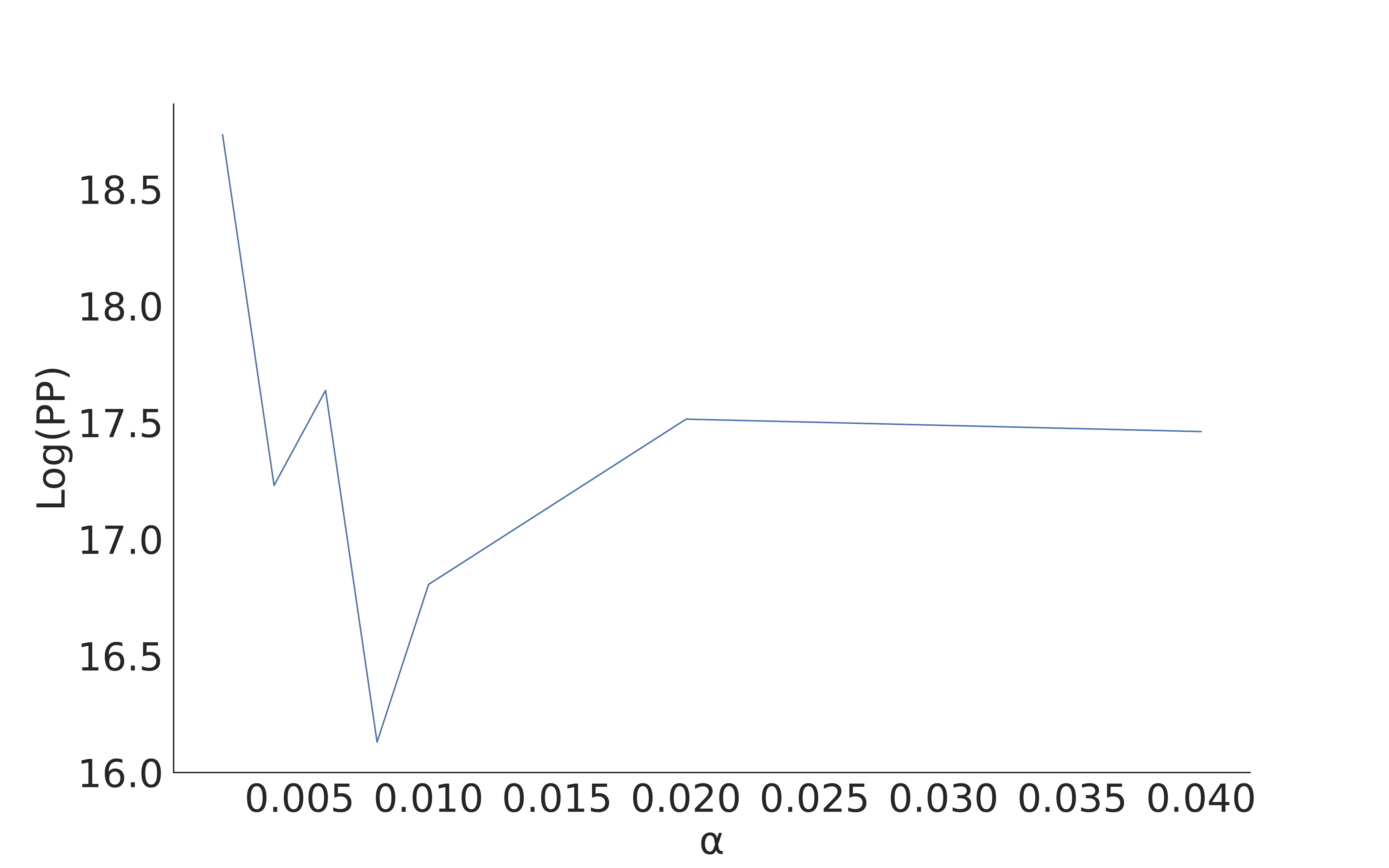}
\end{subfigure}
\caption{Ablation study for the learning rate, i.e., $\alpha$.}
\label{fig:lr}
\end{figure}

In Fig.~\ref{fig:lr}, we ablate the learning rate (i.e., $\alpha$). Since optimization occurs during inference, the number of iterations is fixed, so a higher learning rate ensures convergence. In our experiments, we use $\alpha=0.1$, which has the lowest perplexity, and the highest Retrieval score.  Note that the graphs might be misleading due to the wide range of values. The method is relatively stable to this parameter.

\begin{figure*}[t]
\centering
\begin{subfigure}{1.0\textwidth}
    \includegraphics[width=\textwidth]{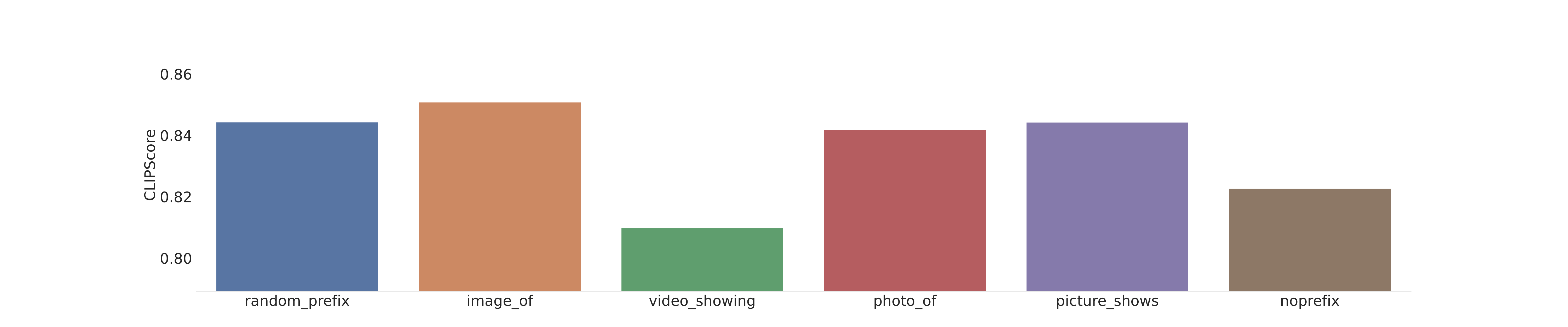}
\end{subfigure}
\hfill
\begin{subfigure}{1.0\textwidth}
    \includegraphics[width=\textwidth]{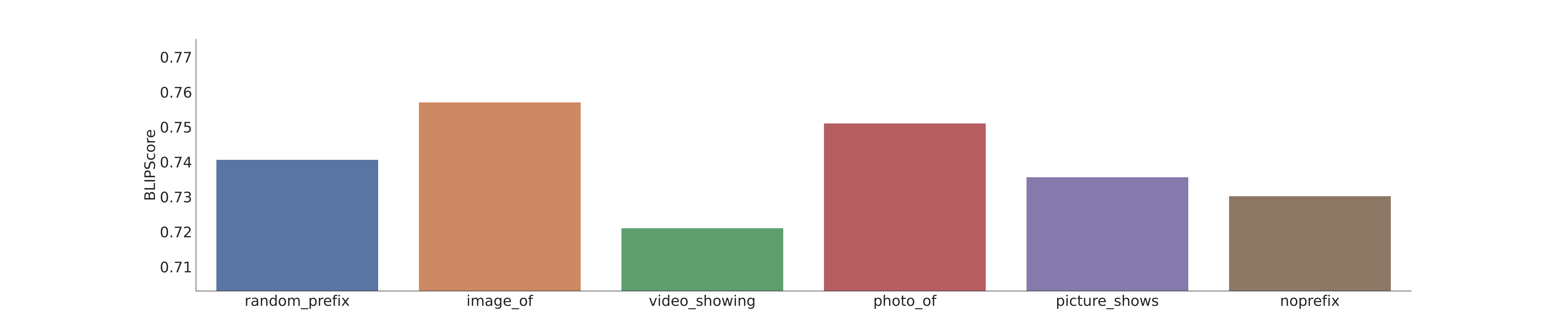}
\end{subfigure}
\hfill
\begin{subfigure}{1.0\textwidth}
    \includegraphics[width=\textwidth]{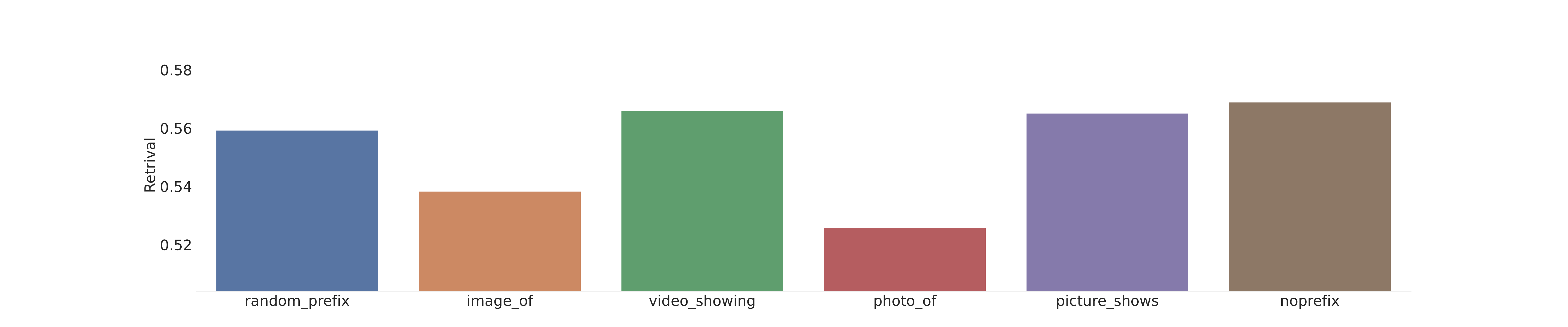}
\end{subfigure}
\hfill
\begin{subfigure}{1.0\textwidth}
    \includegraphics[width=\textwidth]{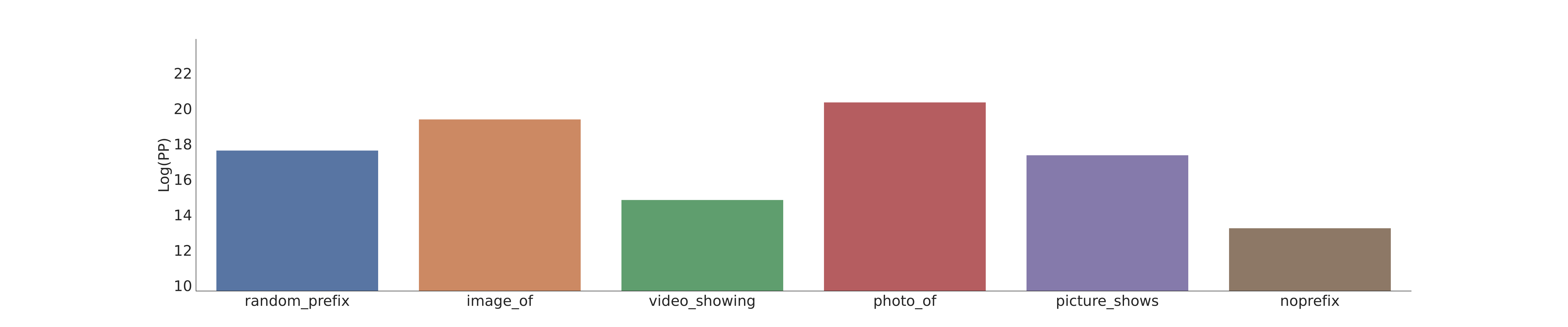}
\end{subfigure}
\caption{Ablation study for different prompts.}
\label{fig:prompts}
\end{figure*}

In Fig.~\ref{fig:prompts}, we study different prompts. In our method, we perturbed a prompt for each generated sentence to increase robustness to different scenarios (e.g., image set captioning and videos). Note that while the option of no prefix at all results in good performance, we find it less focused for the task of visual captioning. 

\begin{figure*}[t]
\centering
\begin{subfigure}{0.47\textwidth}
    \includegraphics[width=\textwidth]{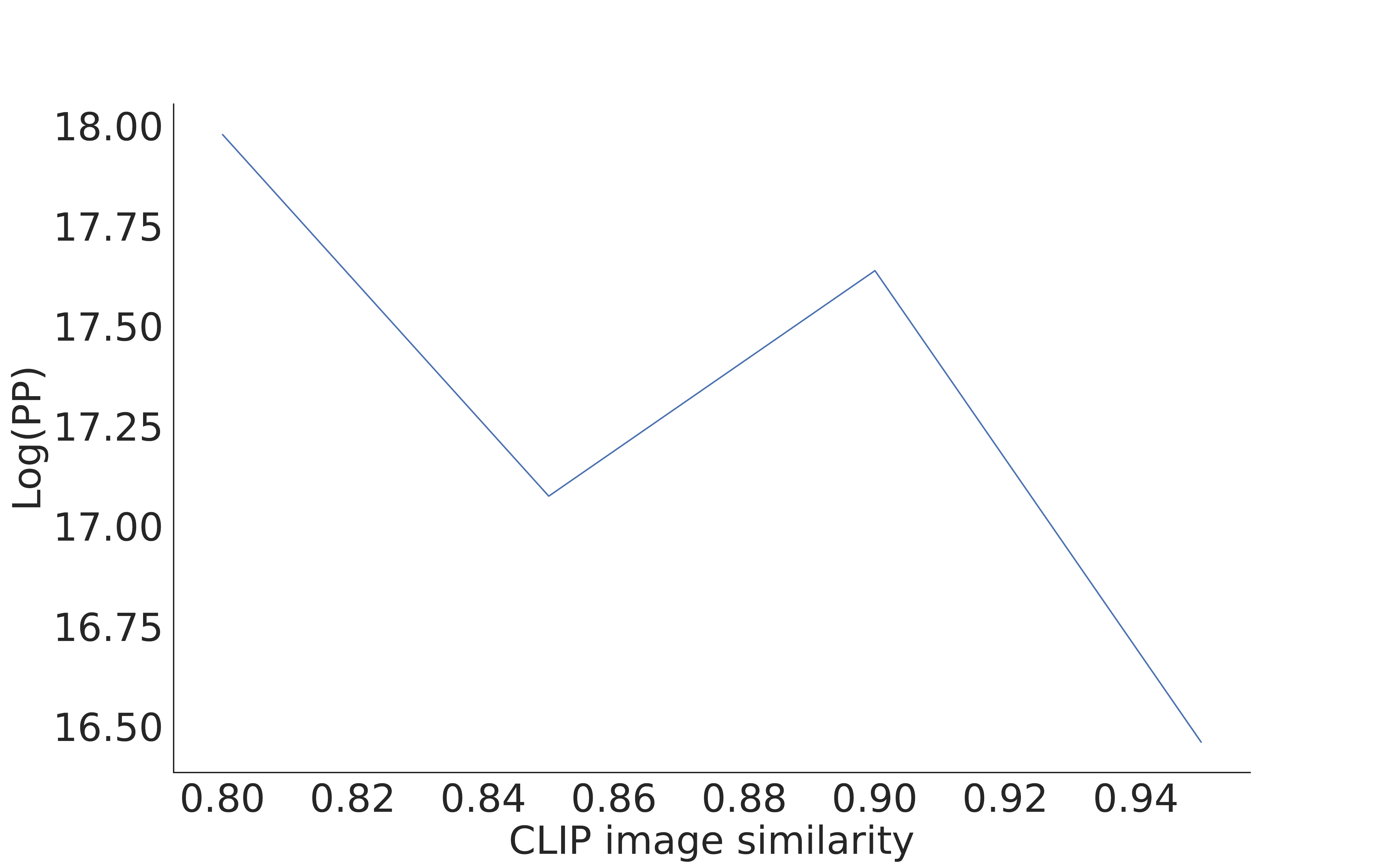}
\end{subfigure}
\hfill
\begin{subfigure}{0.47\textwidth}
    \includegraphics[width=\textwidth]{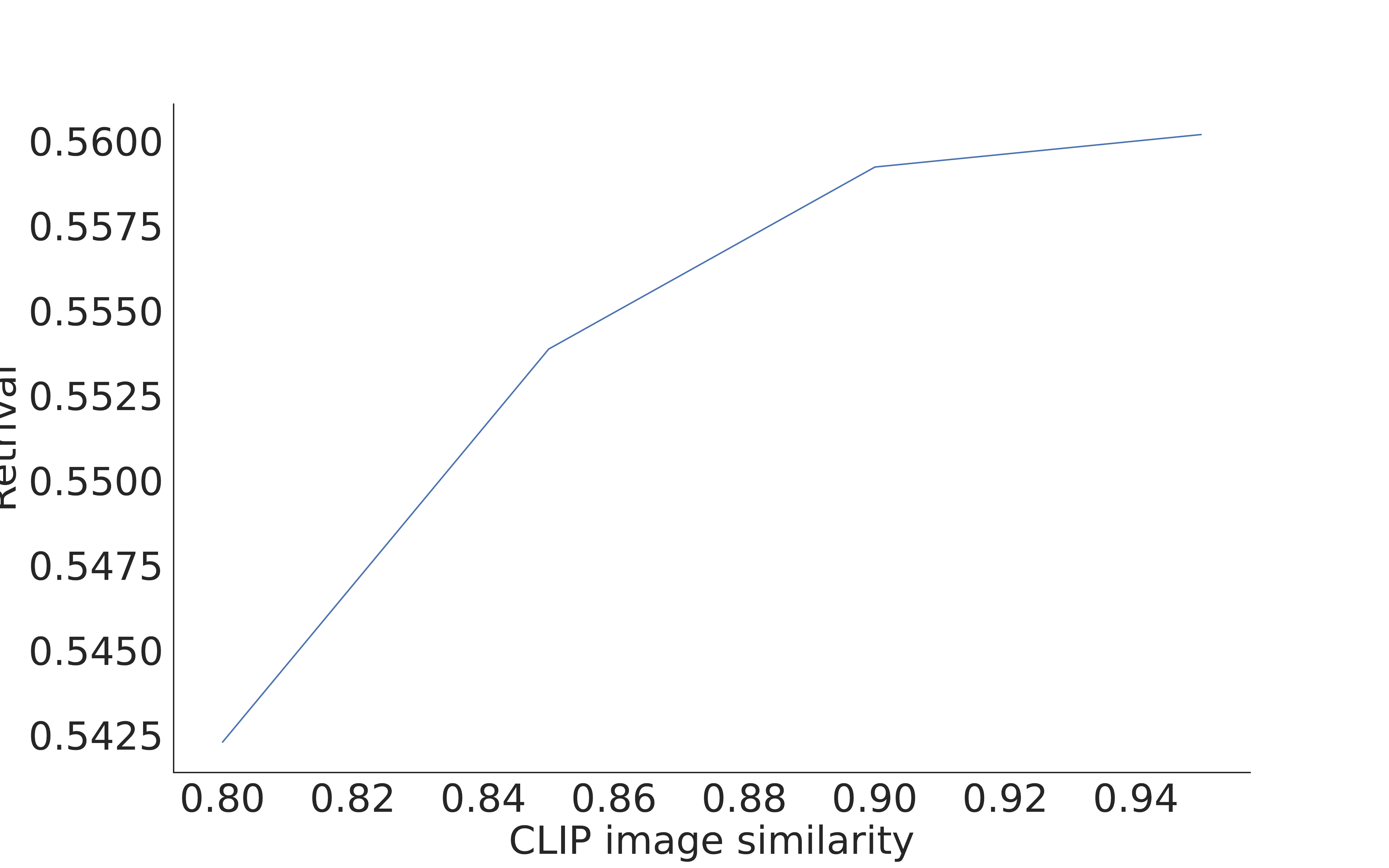}
\end{subfigure}
\caption{Ablation study for the CLIP-based frame selection method. We ablate different threshold values used to pick significant frames.}
\label{fig:abl_frame}
\end{figure*}

In Fig.~\ref{fig:abl_frame}, we assess our CLIP-based sampling method. Our method employs CLIP's visual encoder to compute image similarity. The Retrieval score increases, as can be expected, with the CLIP image similarity. The perplexity score is relatively stable, but there is a trade-off between the two. %is stable at a large range of values, while the CLIP similarity is, as expectedsaturates at value of $0.9$, which is the threshold we set. 

\begin{figure}[t]
\centering
\begin{subfigure}{0.47\textwidth}
    \includegraphics[width=\textwidth]{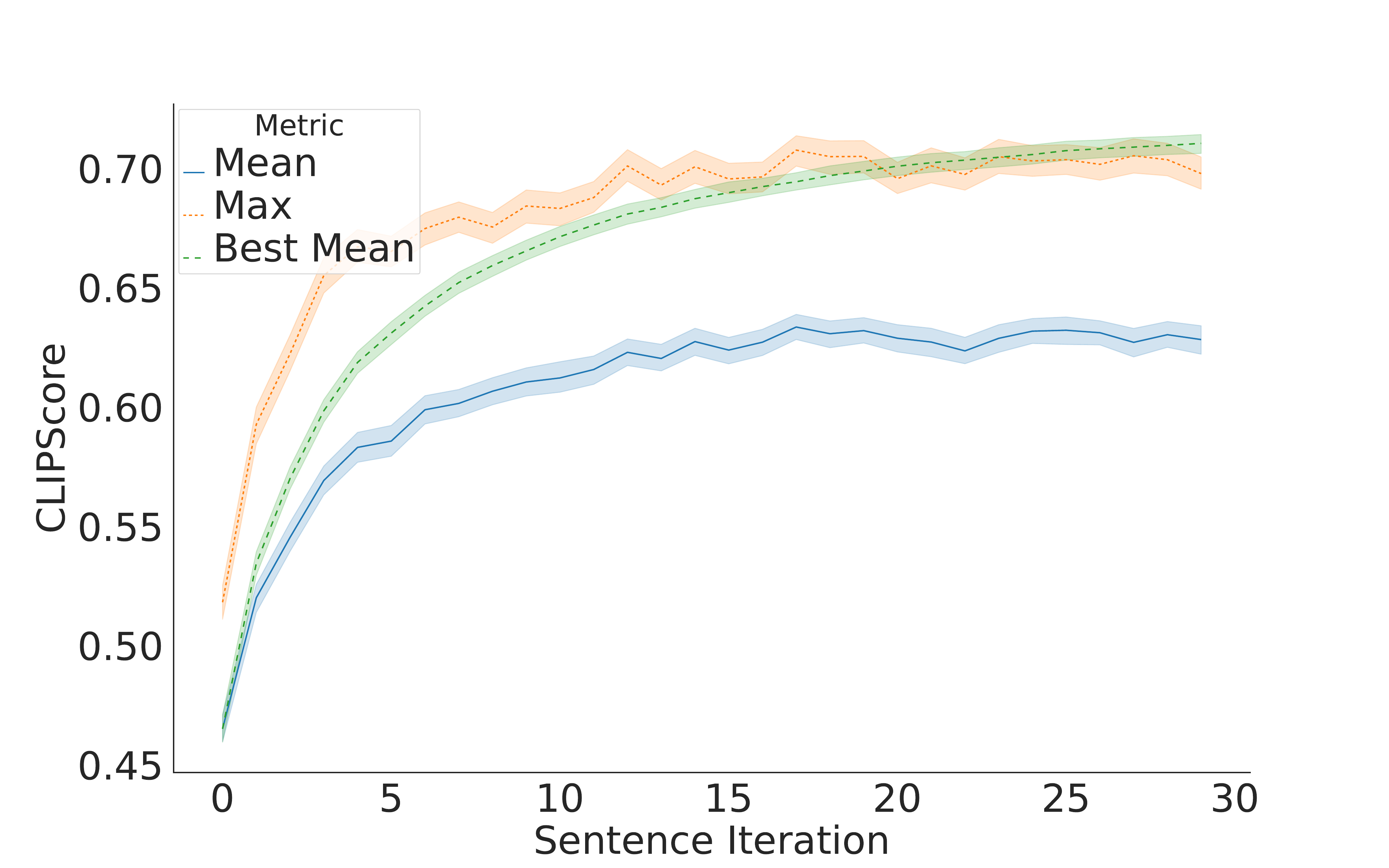}
    \caption{Two images}
\end{subfigure}
\hfill
\begin{subfigure}{0.47\textwidth}
    \includegraphics[width=\textwidth]{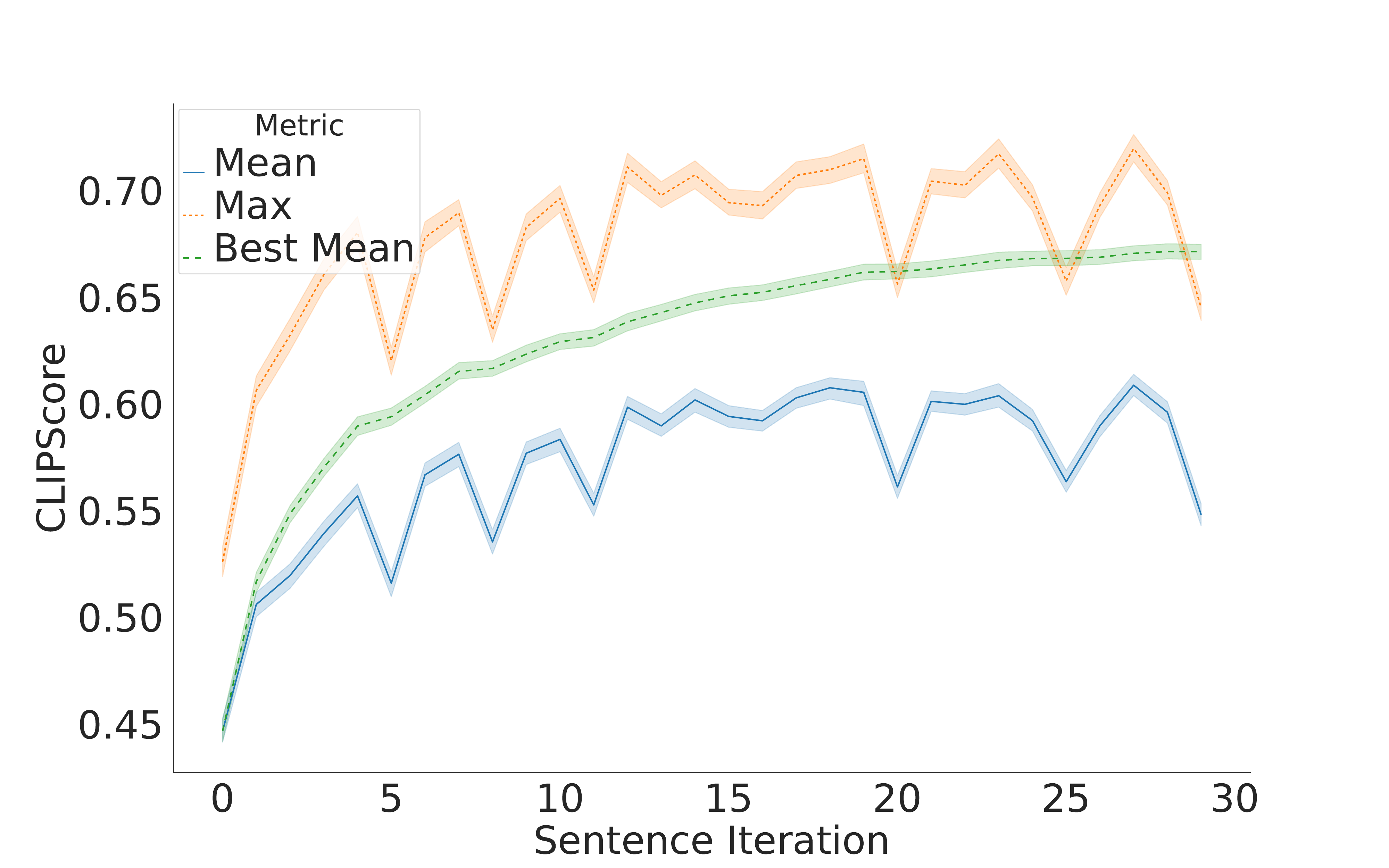}
    \caption{Three images}
\end{subfigure}
\hfill
\begin{subfigure}{0.47\textwidth}
    \includegraphics[width=\textwidth]{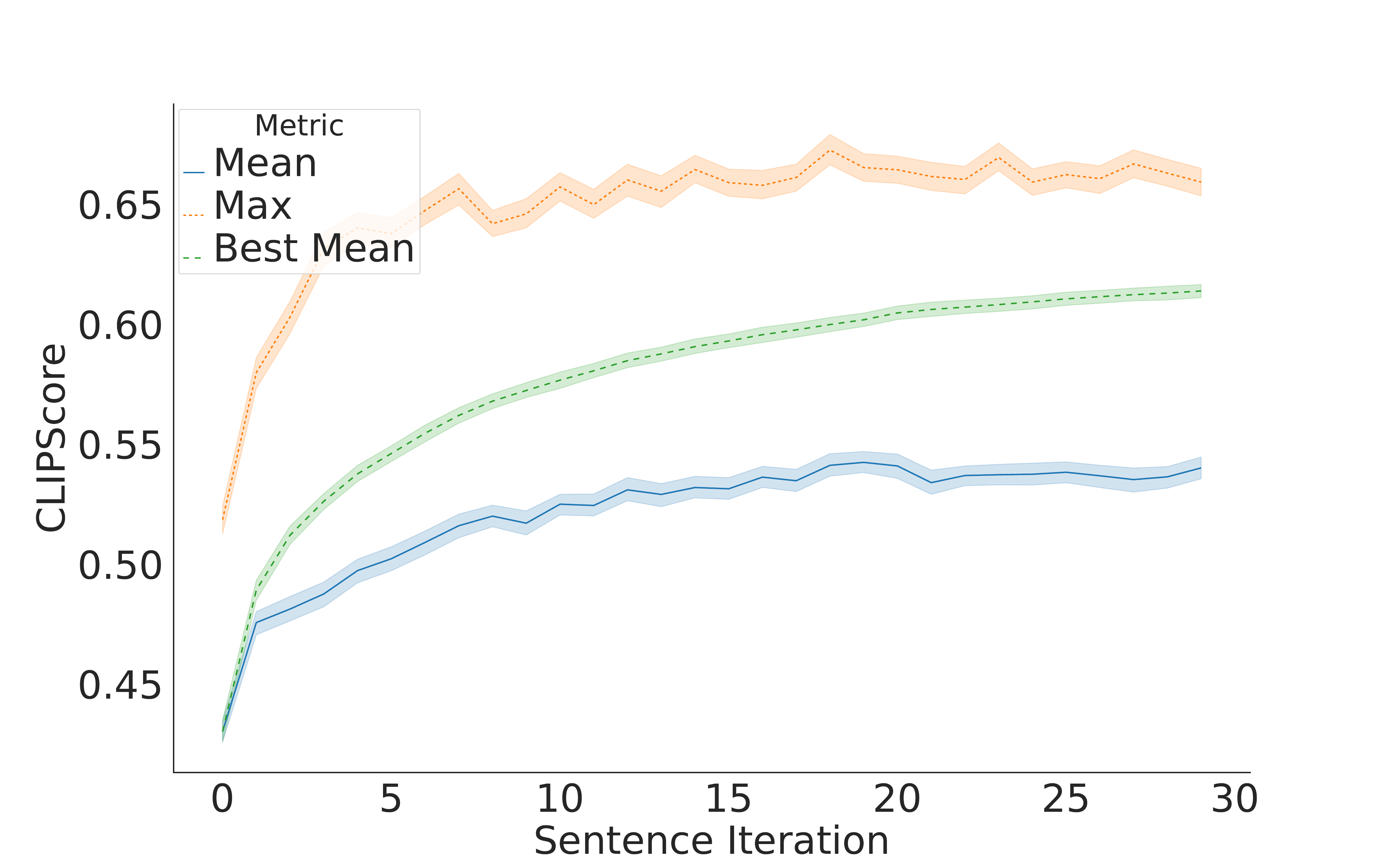}
    \caption{Four images}
\end{subfigure}
\caption{CLIP Score progress over the generation process.}
\label{fig:clip_iter}
\end{figure}

In Fig.~\ref{fig:clip_iter}, we demonstrate how the CLIP score progresses during the generation process. We report the following statistics: (i) Mean is the average CLIP score at the given iteration across the set (ii) Max is the maximum CLIP score at the given iteration across the set (iii) Best Mean is the best mean score up to this iteration. The challenge of fitting to multiple visual cues can cause performance instability during optimization. Thus, we suggest selecting the sentence with the highest CLIP score from all the generated sentences.  

\section{Qualitative Analysis}
\begin{figure*}[t]
	\centering
    \includegraphics[width=1\linewidth]{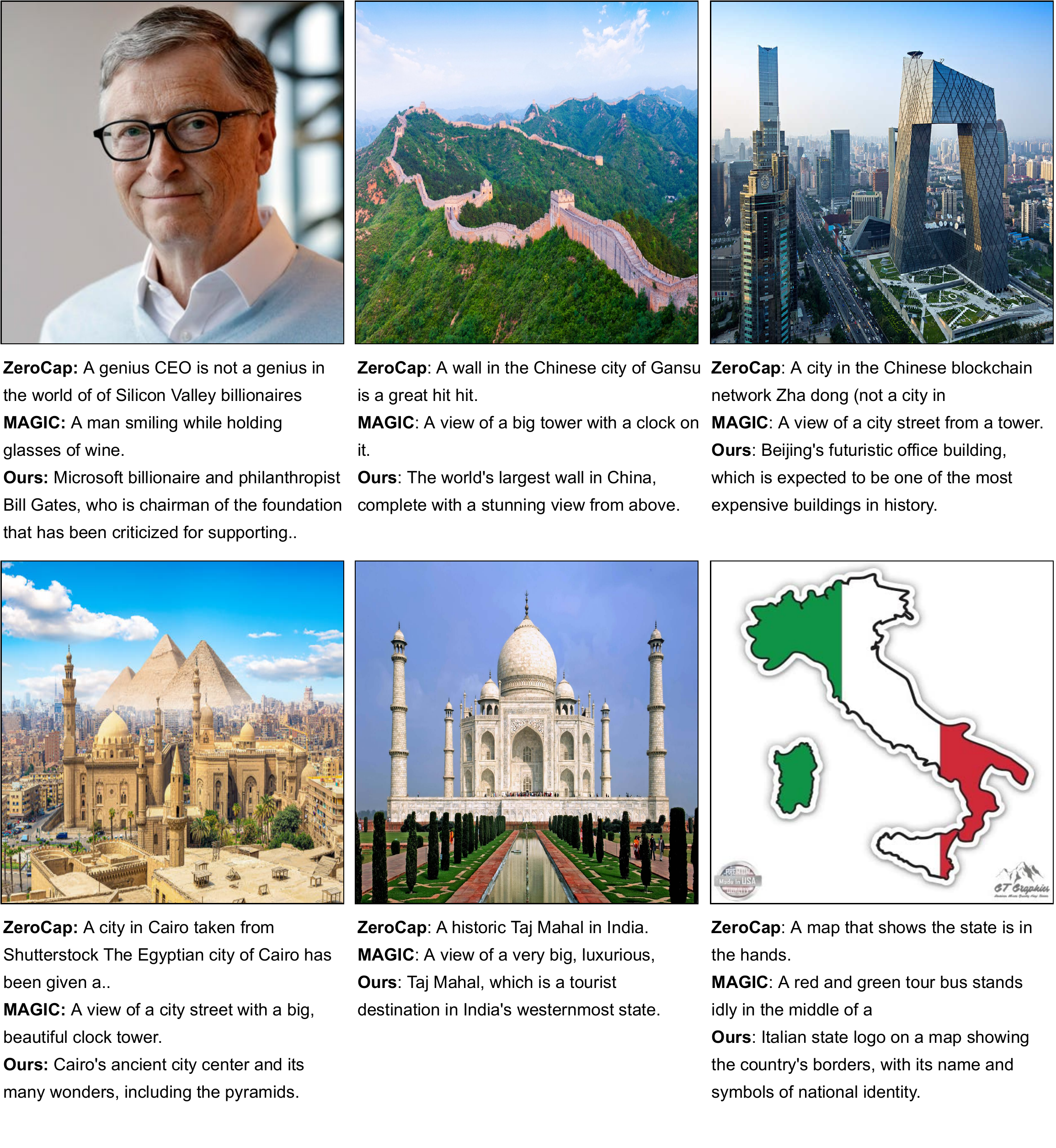}
    \caption{Examples of our image captions on examples that require real-world knowledge, with two zero-shot image captioning baselines.}
    \label{fig:real_world_image}
\end{figure*}

In Fig.~\ref{fig:real_world_image} we demonstrate the ability of our model to caption images while incorporating real-world knowledge. We provide comparison with two baseline zero-shot image captioning models: ZeroCap~\cite{tewel2021zero} and MAGIC~\cite{2022arXiv220502655S}. The baseline models fail to describe the images with fluent natural language, nor do they provide real-world information. Our model recognizes Bill Gates, the Great Wall of China and the map of Italy, and incorporates that knowledge in rich descriptive captions.

\begin{figure*}[t]
	\centering
    \includegraphics[width=1\linewidth]{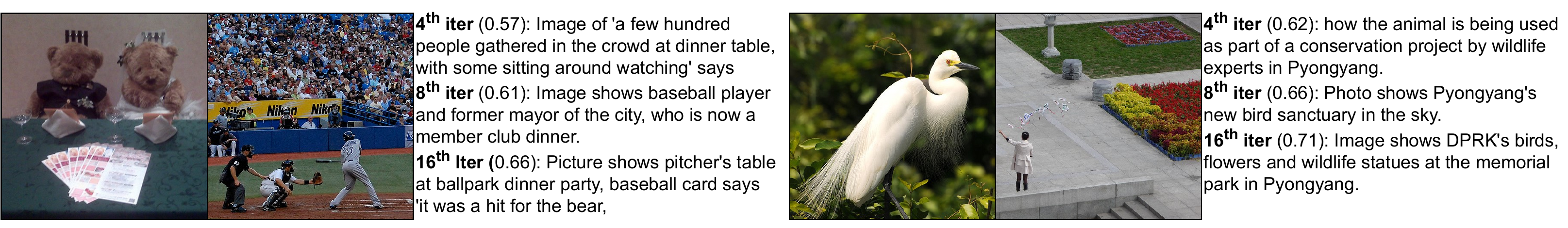}
    \caption{Evolution of image pair captions. We show the sentence with the highest CLIP score at different generation iterations.} 
    \label{fig:evol_cap}
%\end{figure*}
\medskip\smallskip
%\begin{figure*}[t]
	\centering
    \includegraphics[width=1\linewidth]{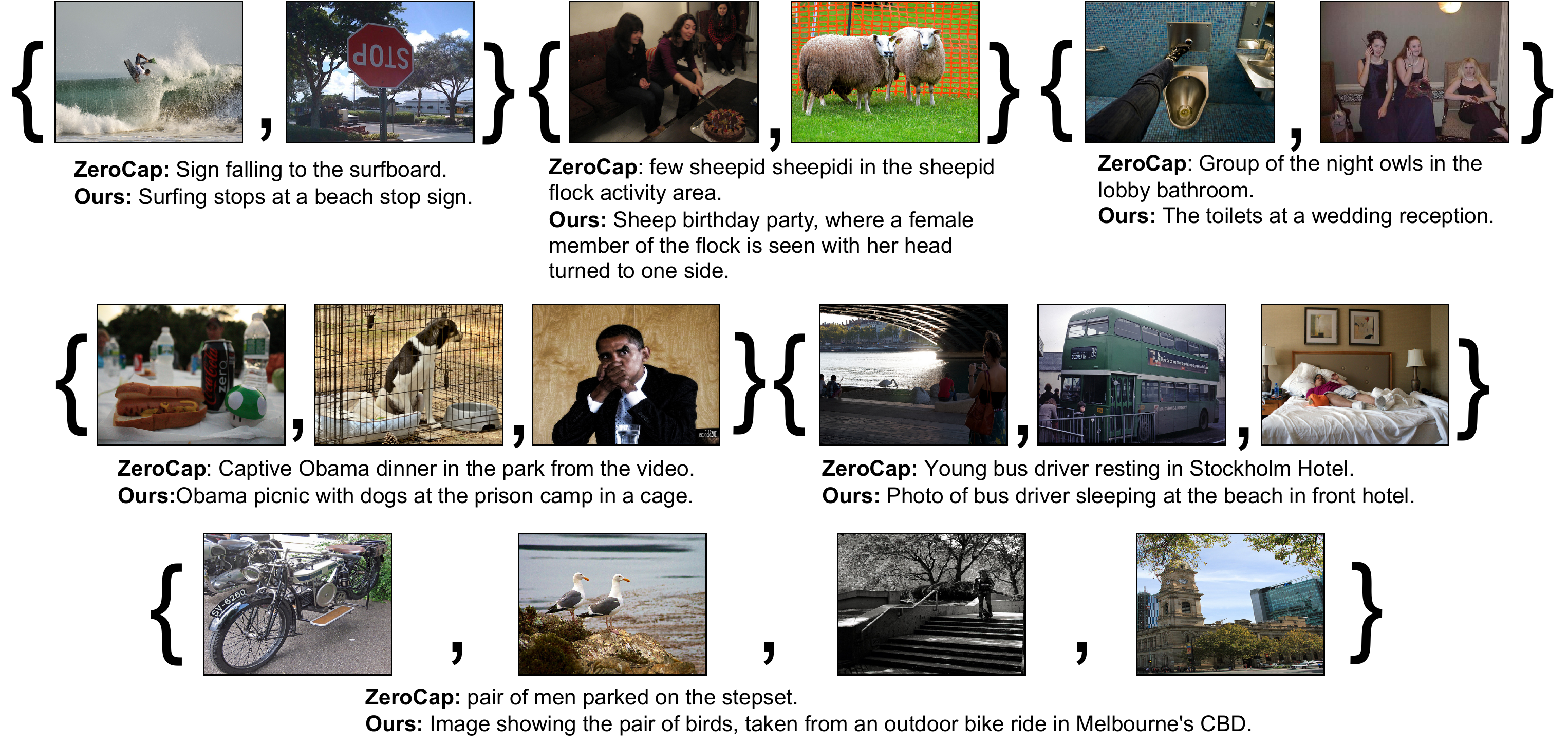}
    \caption{Examples of our image set captioning, for different set sizes. We compare our method with  ZeroCap, another zero-shot method.} 
    \label{fig:sets_qualitative}
\end{figure*}

In Fig.~\ref{fig:evol_cap} we illustrate the progression of captions during the optimization process. The left image set shows two very different pictures of a toy bear and a baseball game. Earlier captions discuss the crowd and the dinner separately. The eighth iteration improves grounding, and the method recognizes the baseball game. A coherent narrative is built in the 16th iteration. It is described as a table of a pitcher at a dinner party. There are baseball cards on the table, and bears serve as a metaphor for phrasing a quote. For the right image set, after four iterations our method generates a caption that includes the word Pyongyang as the location and the word 'wildlife'. At the eighth iteration, the caption identifies the animal as a bird. As a result of detecting Pyongyang as the location, the bird is described as being from the DPRK. A reference is also made to the flowers.

 In Fig.~\ref{fig:sets_qualitative} we illustrate captions generated for sets of various sizes. We are able to identify and describe the content of two images even if there is no significant correlation between them. A stop sign and surfing images are translated to ``Surfing stops...", while pictures of a toilet and ladies in formal attire are captioned with ``The toilets at the wedding reception.''. Also, pictures of a sheep and a birthday cake are captioned with ``Sheep's birthday...''.
 
 When three images can be described with a coherent story, the model can do so. As an example, for a set of images of a bus, a hotel bed, and a beach, our method generates the caption: ``Photo of bus driver sleeping on the beach from the hotel.''. This caption grounds all the images while still creating a plausible narrative. In addition, even when real-world knowledge is necessary, e.g., a picture of Obama, the caption relates to it. 
 
 Our method was able to produce a coherent narrative even when dealing with a complex case of four images. It describes a narrative of an image of birds taken while cycling in Melbourne. We note that ZeroCap's sentences tends to create an irrational context, e.g., ``Captive Obama...,'' which is perhaps the result of token-based optimization rather than sentence-based optimization.

\begin{figure*}[t]
	\centering
    \includegraphics[width=1\linewidth]{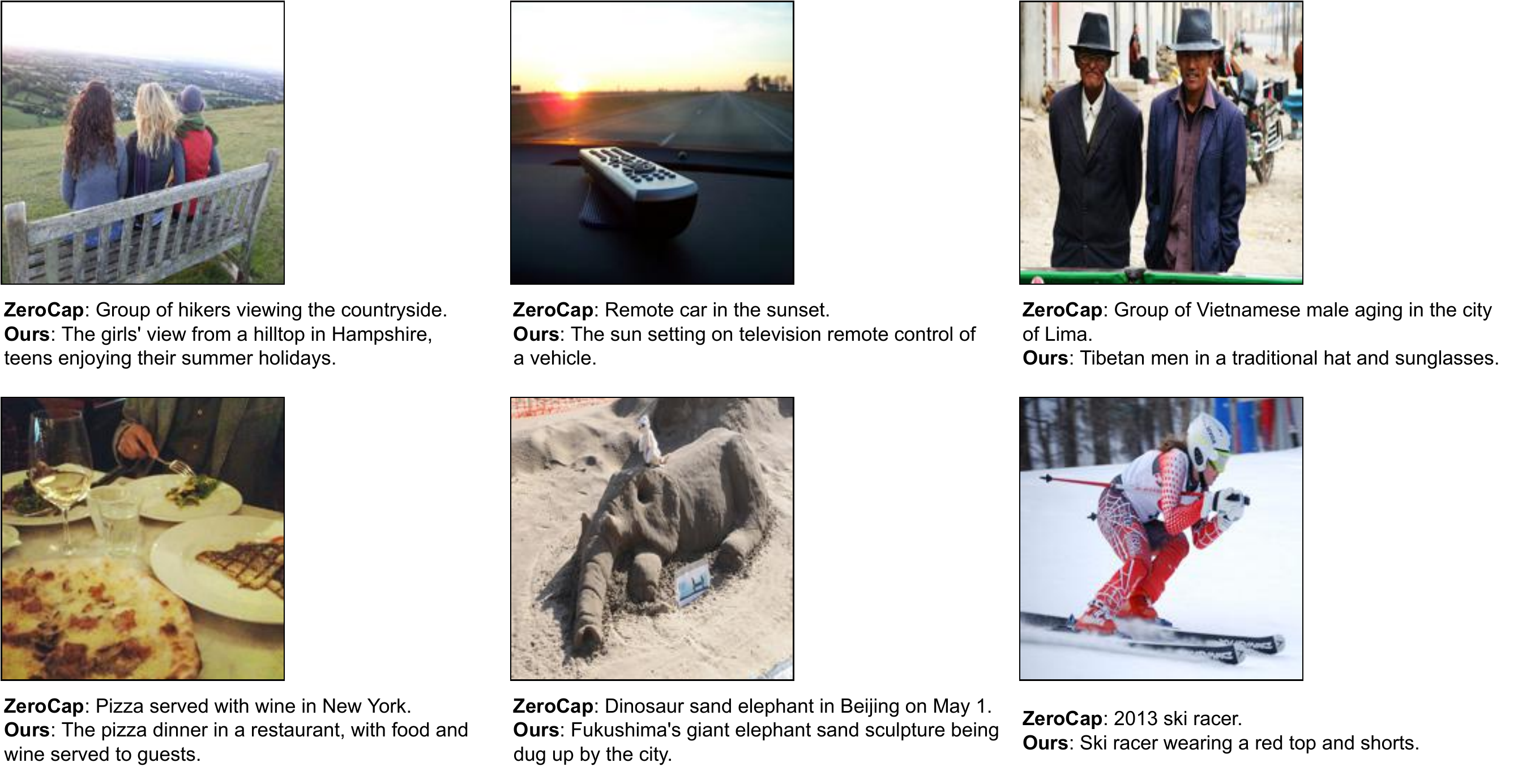}
    \caption{Examples of image captions.}
    \label{fig:image_caption}
\end{figure*}

In Fig.~\ref{fig:image_caption}, we demonstrate our method's zero-shot image captioning capabilities. We compare with ZeroCap. We find that ZeroCap's captions are more direct, whereas the narrative of our captions is more natural. For example, in the first row, on the left, the captions describe the girls and indicate that it is their summer vacation, whereas ZeroCap mentions what appears in the image. These results might come from the way we construct sentences. By letting the PLM construct sentences, we improve language fluency. ZeroCap, on the other hand, alternates each token to correspond to the image, which might hinder the language.  

\begin{figure*}[t]
	\centering
    \includegraphics[width=1\linewidth]{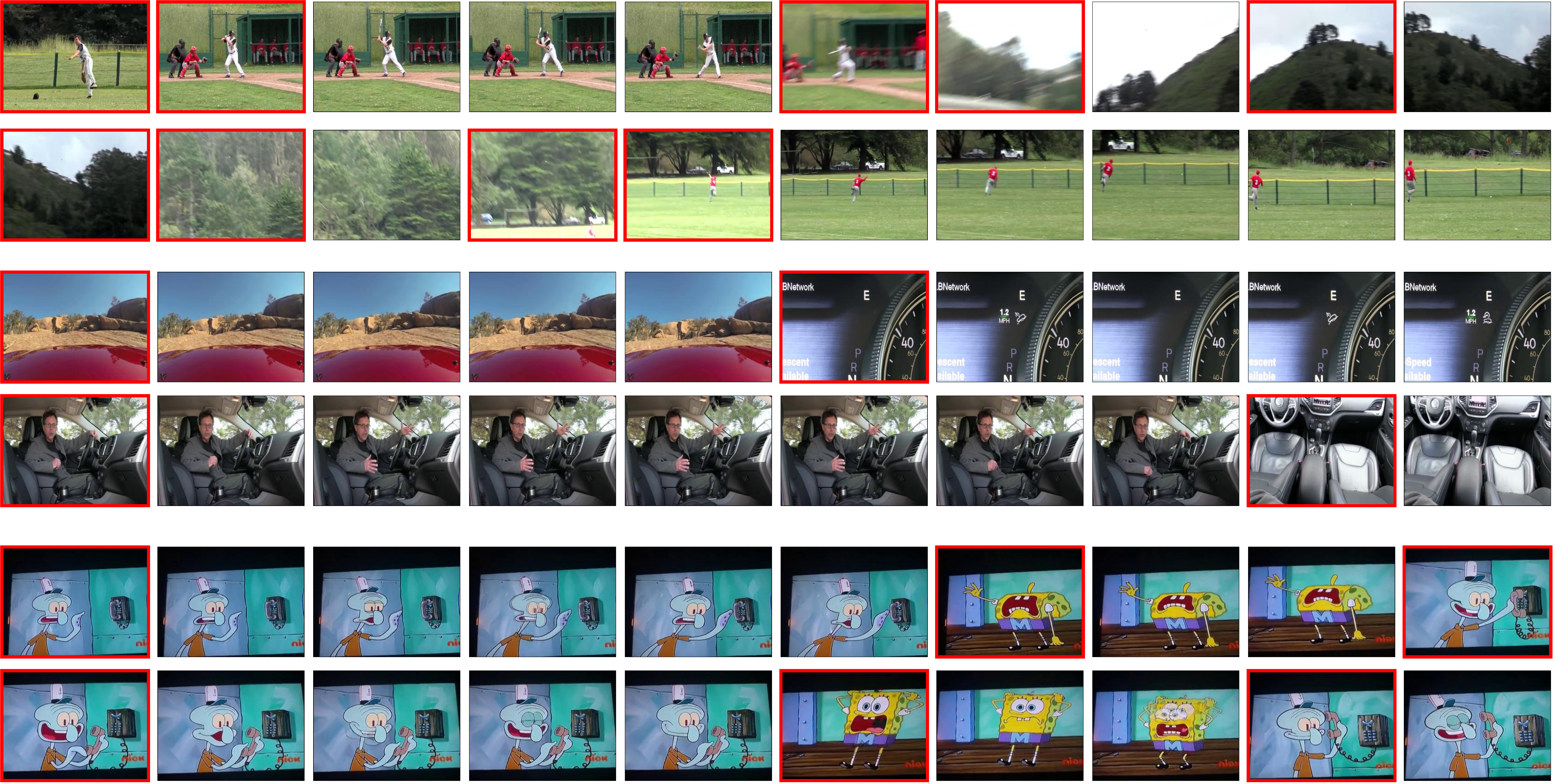}
    \caption{Illustration of our CLIP-based sampling strategy. The picked frames are outlined in red.}
    \label{fig:picked_frames}
\end{figure*}

In Fig.~\ref{fig:picked_frames}, we illustrate the CLIP-based mechanism we use to pick novel and diverse frames. As a result of using CLIP image similarity, the method is able to find frames with a very different content, e.g. where the environment or objects change. We highlight the selected frames with a red border. For example, the first row contains a frame depicting a pitcher, followed by a frame showing the catcher. Frames following this one are ignored until the ball is hit. In the following video, only four frames are selected, filtering out many repetitions. The strategy also works with animations, as shown in the third video.

In Fig.~\ref{fig:qual_video_supp}, we show more examples for the full generation process for videos. We present the frames selected by our CLIP-based sampling method for each video. Additionally, we report BERT-based perplexity score and CLIP score. The low perplexity score indicates that early sentences have good language, but subsequent sentences improve the CLIP score significantly. Our method can ground objects and generate coherent sentences in various contexts. 

In Fig.~\ref{fig:sets2}, we illustrate the evolution of sentences, using two images. Interestingly, the method uses stories to weave the photos into a coherent story. For instance, the image of prison and a bedroom photo results in a caption about a prisoner's bedroom. 

In Fig.~\ref{fig:sets3}, three images are employed, and the generation process is displayed. Often, creating a coherent sentence from three images is too challenging. Therefore, in those cases, it is better to choose the sentence based on the perplexity indicator rather than using the CLIP score. Thus, the language will be fluent as it describes a story line without describing everything in every image. Fig~\ref{fig:sets4} shows the same phenomenon when there are four images.

\clearpage

\begin{figure*}[t]
	\centering
    \includegraphics[width=1\linewidth]{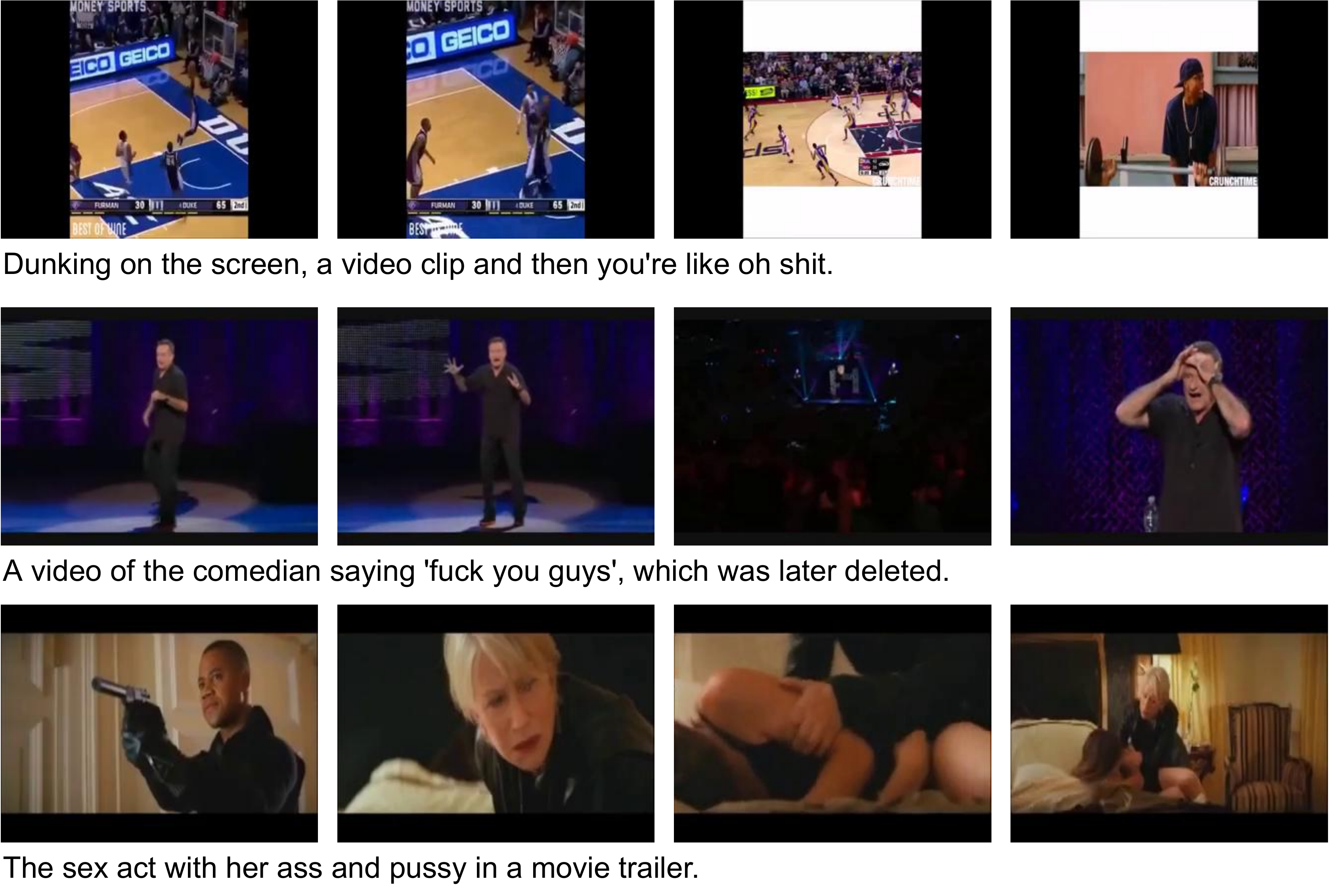}
    \caption{Example of harsh language being generated by our model. This illustrates a limitation of web-scale models.}
    \label{fig:toxicity}
\end{figure*}

\begin{figure*}[t]
	\centering
	\caption{The evolution of captions for videos. (below and for multiple pages)}
    \includegraphics[width=1\linewidth]{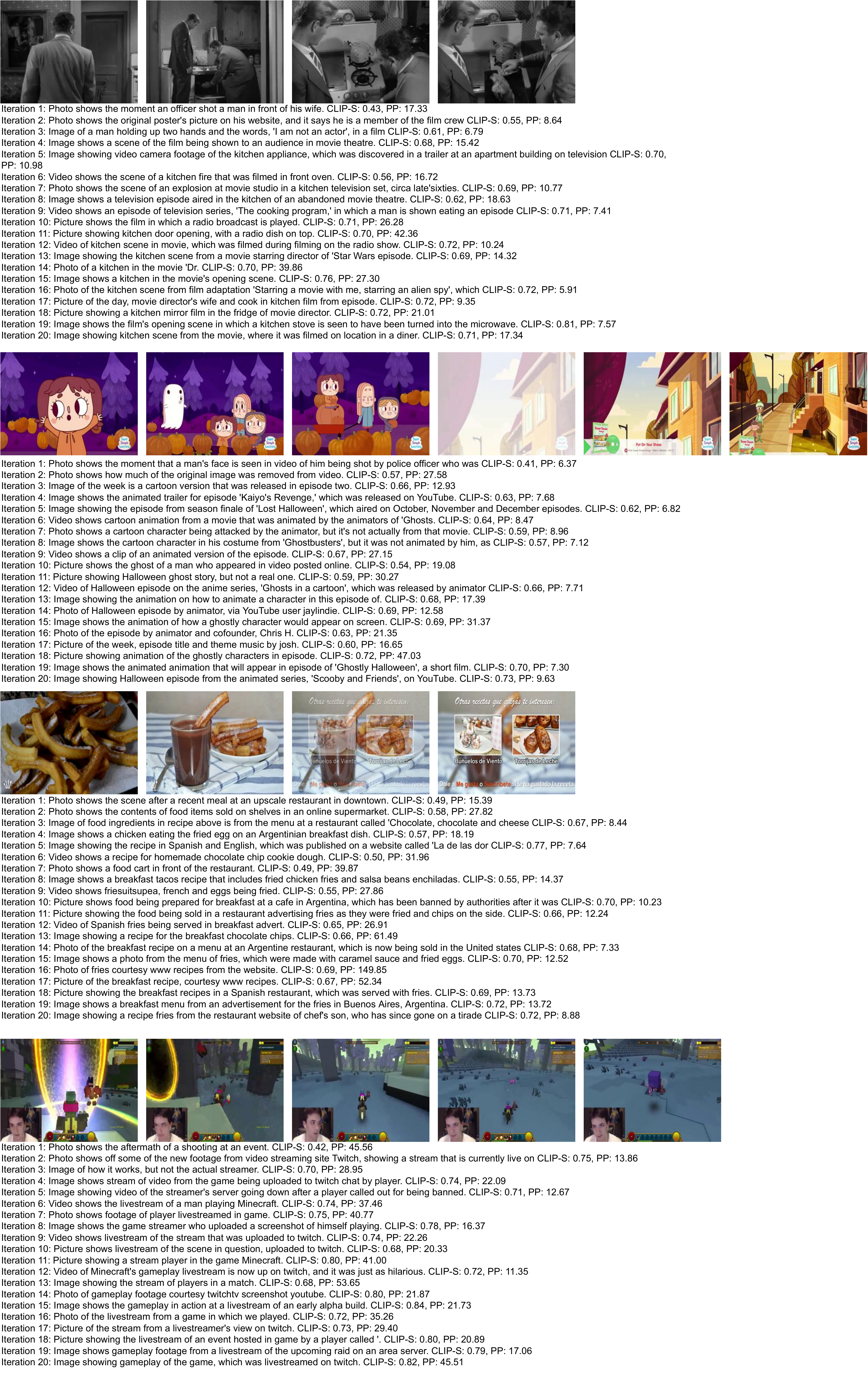}
    \label{fig:qual_video_supp}
\end{figure*}
\begin{figure*}[h!]
	\centering
    \ContinuedFloat
    \includegraphics[width=1\linewidth]{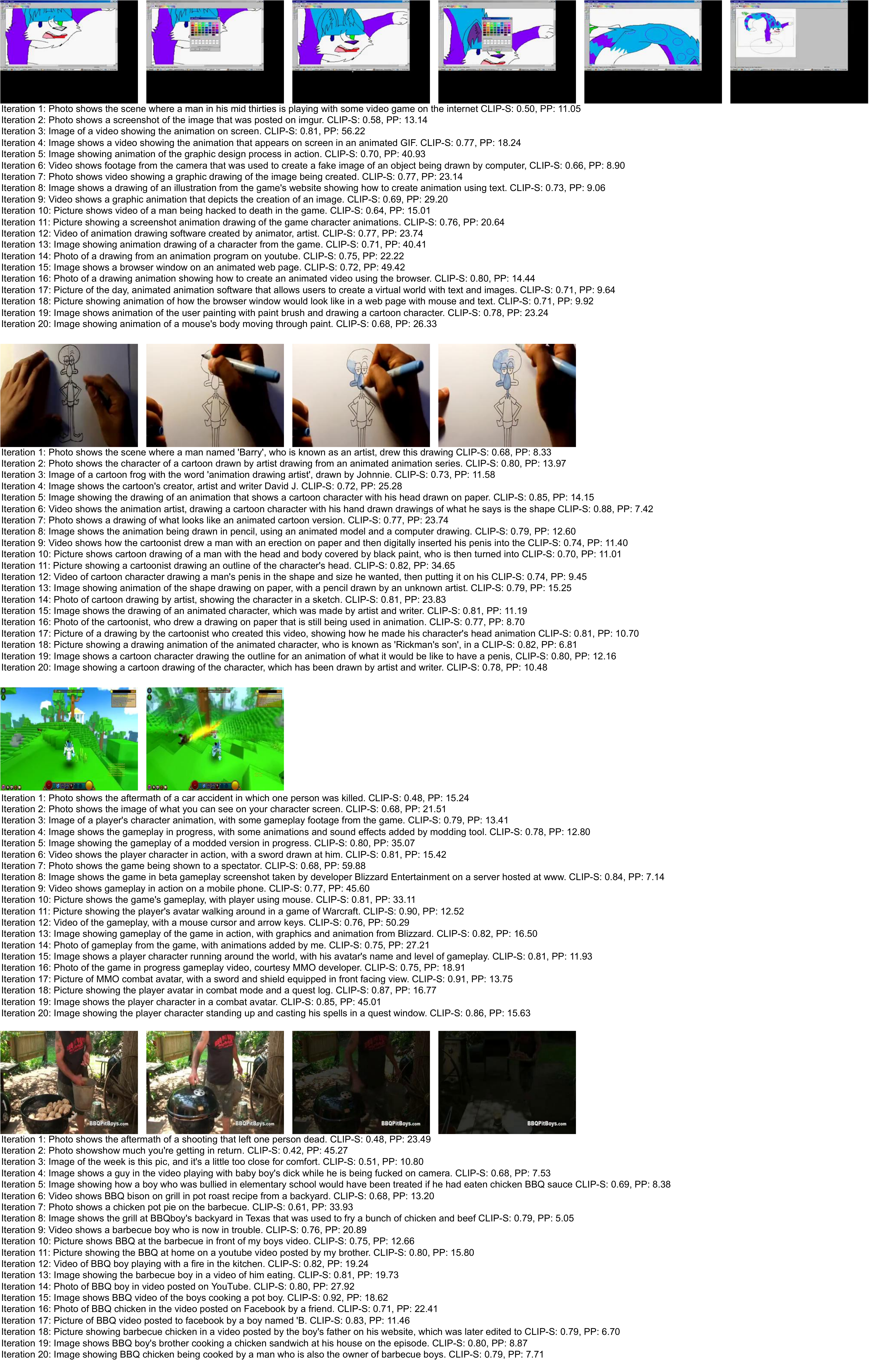}
\end{figure*}
\begin{figure*}[h!]
	\centering
    \ContinuedFloat
    \includegraphics[width=1\linewidth]{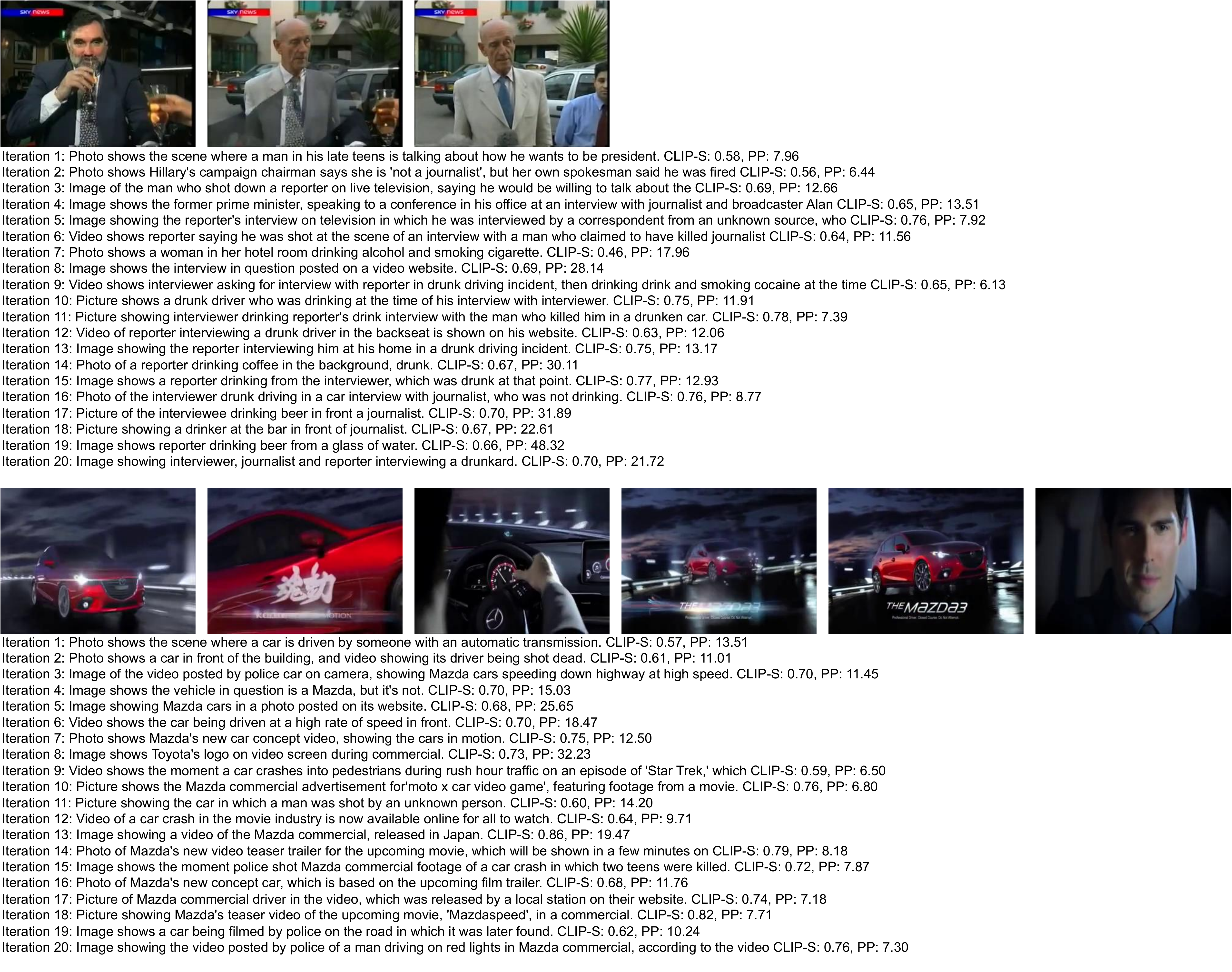}
\end{figure*}

\begin{figure*}[t]
	\centering
	\caption{The evolution of captions for two images in an image set. (below and for multiple pages)}
    \includegraphics[width=1\linewidth]{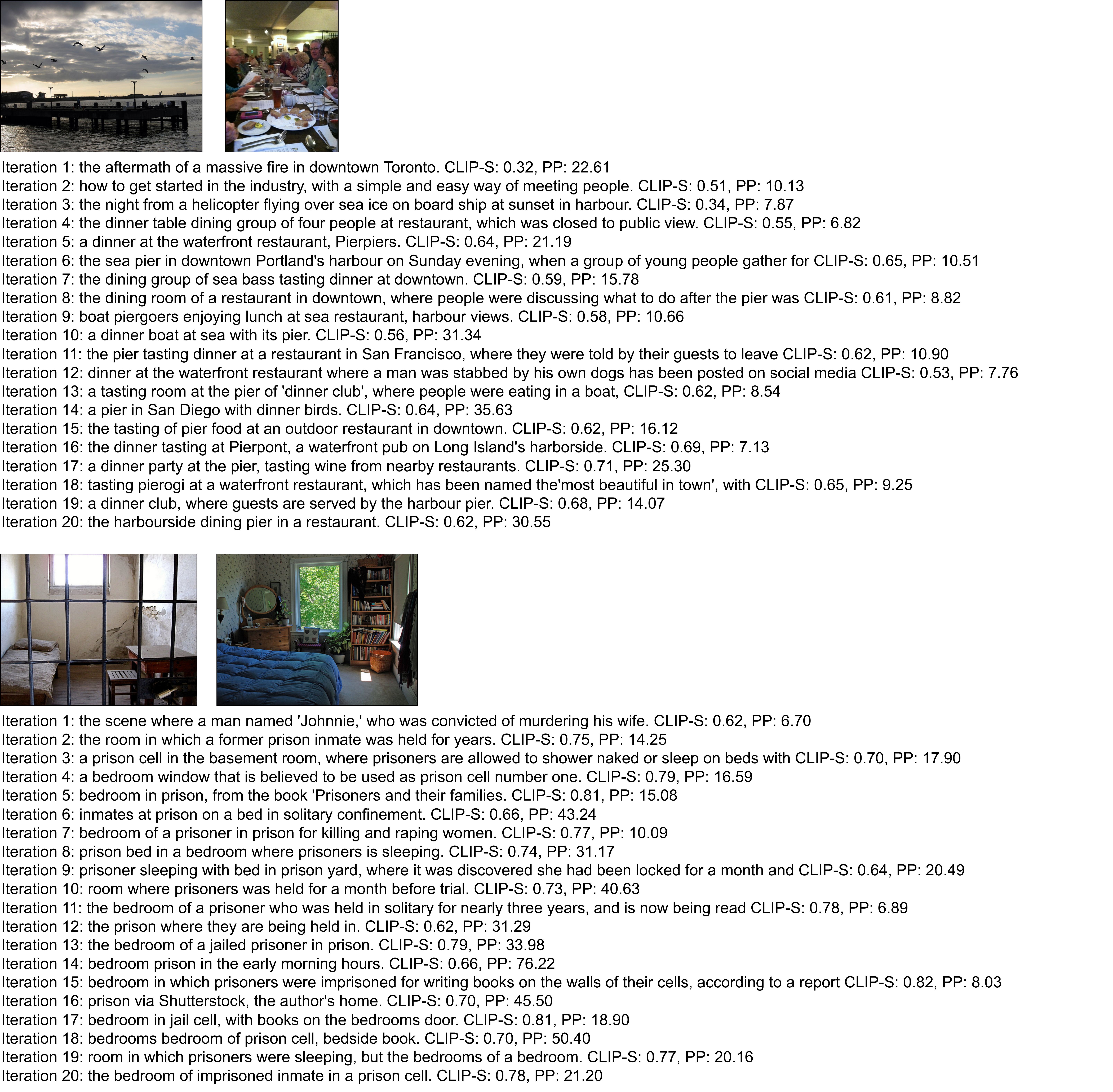}
    \label{fig:sets2}
\end{figure*}
\begin{figure*}[h!]
	\centering
    \ContinuedFloat
    \includegraphics[width=1\linewidth]{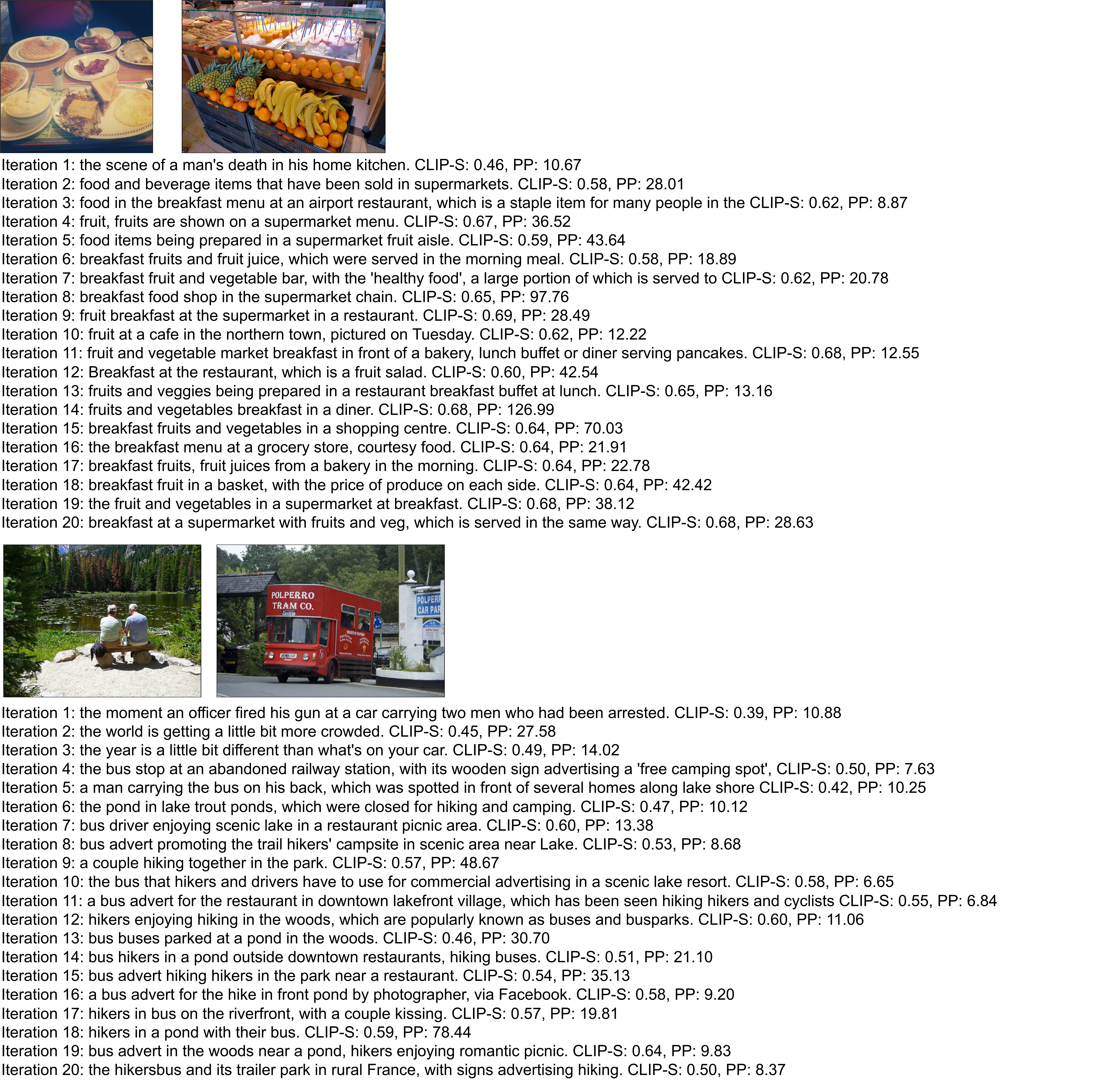}
\end{figure*}

\begin{figure*}[t]
	\centering
	\caption{The evolution of captions for three images in an image set.(below and for multiple pages)}
    \includegraphics[width=1\linewidth]{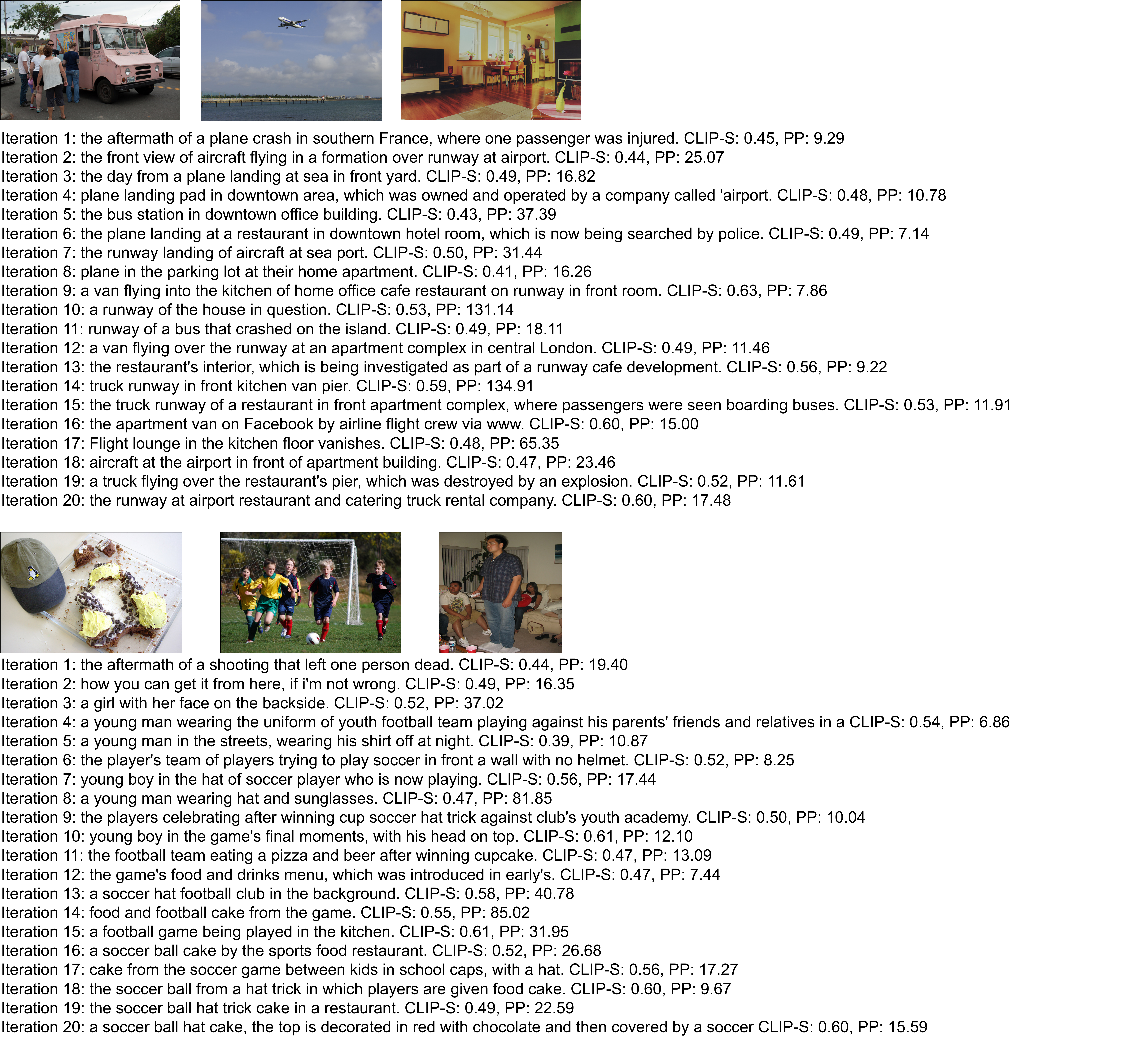}
    \label{fig:sets3}
\end{figure*}
\begin{figure*}[h!]
	\centering
    \ContinuedFloat
    \includegraphics[width=1\linewidth]{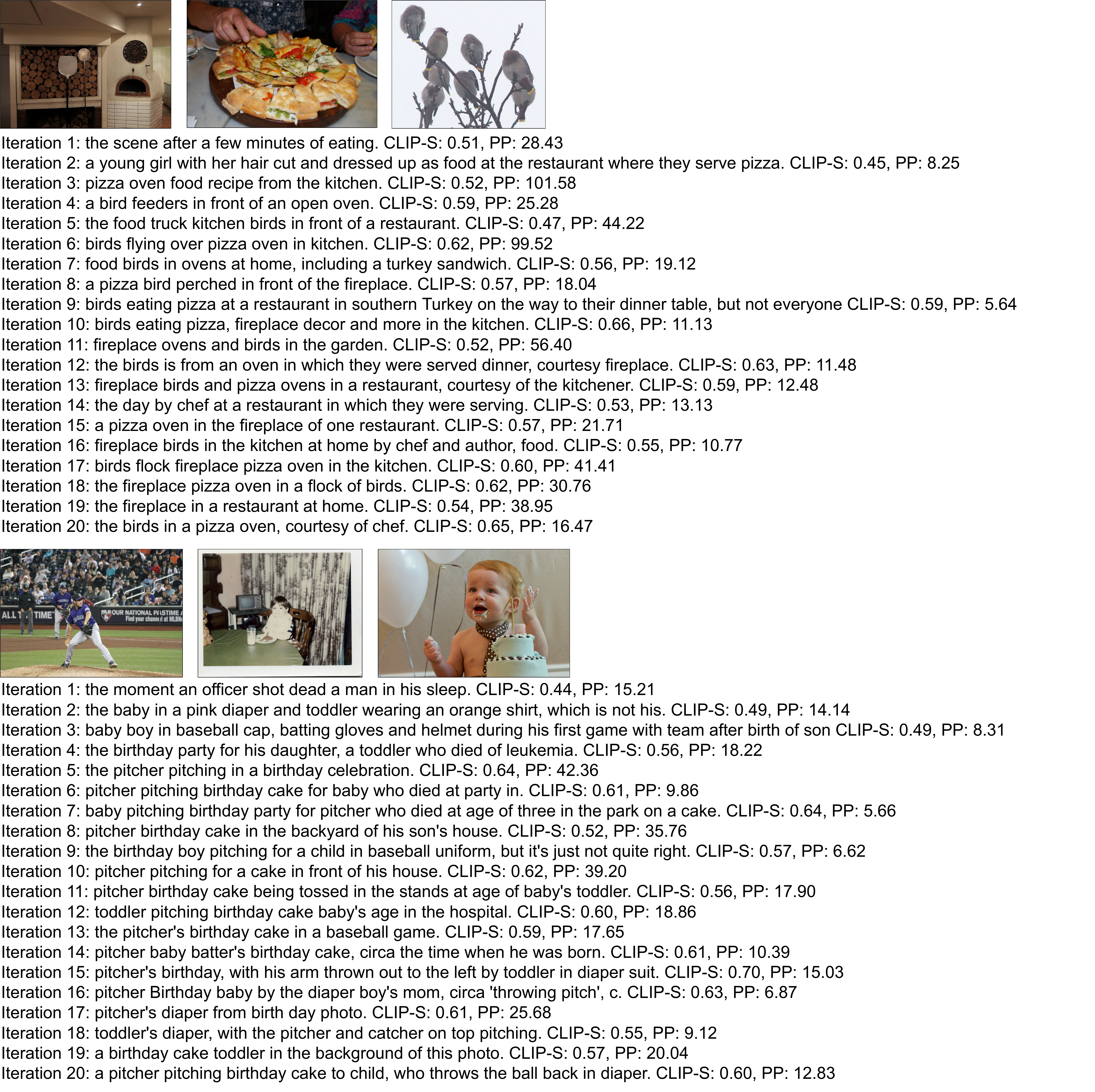}
\end{figure*}

\begin{figure*}[t]
	\centering
	\caption{The evolution of captions for four images in an image set. (below and for multiple pages)}
    \includegraphics[width=1\linewidth]{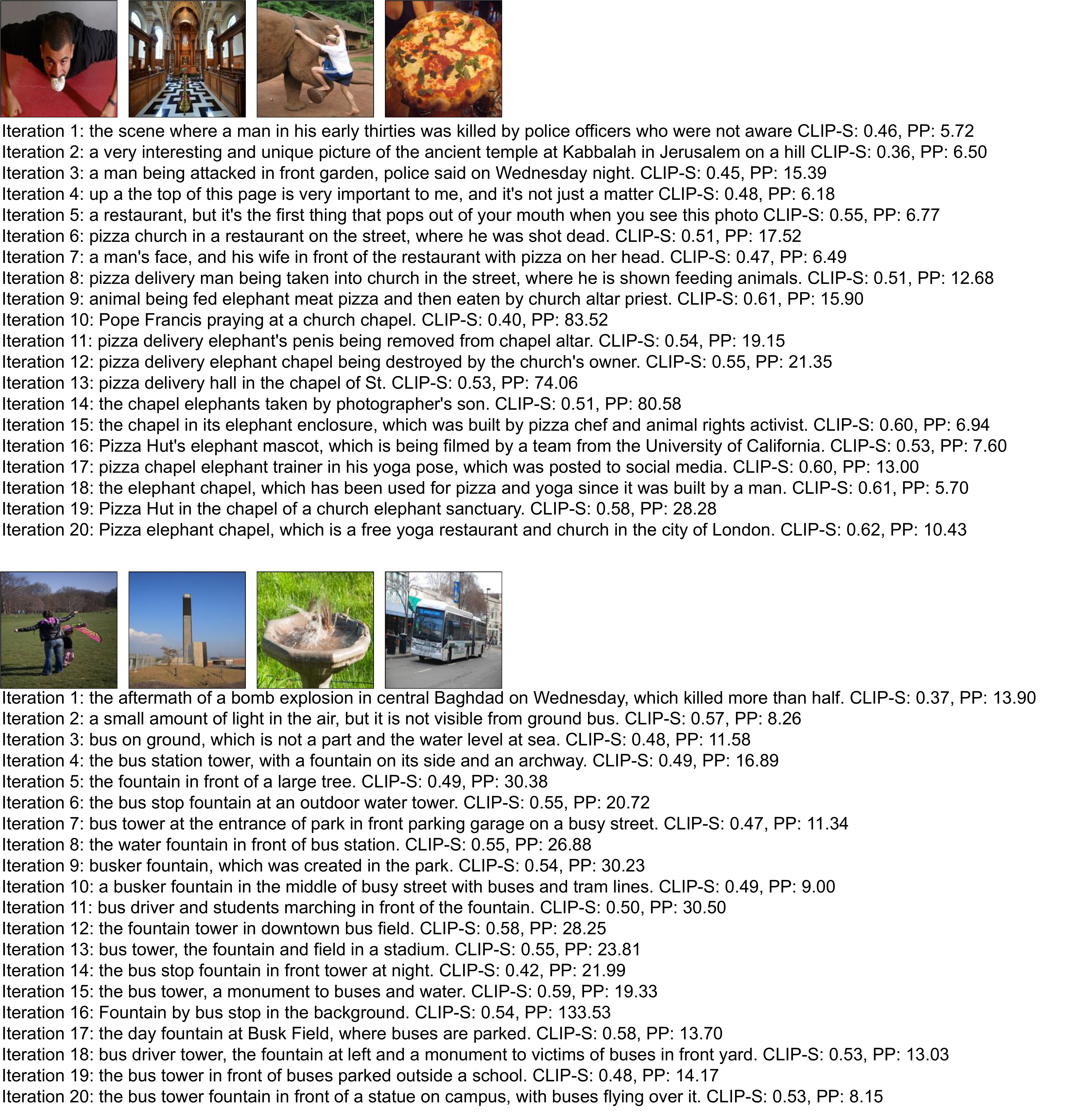}
    \label{fig:sets4}
\end{figure*}
\begin{figure*}[h!]
	\centering
    \ContinuedFloat
    \includegraphics[width=1\linewidth]{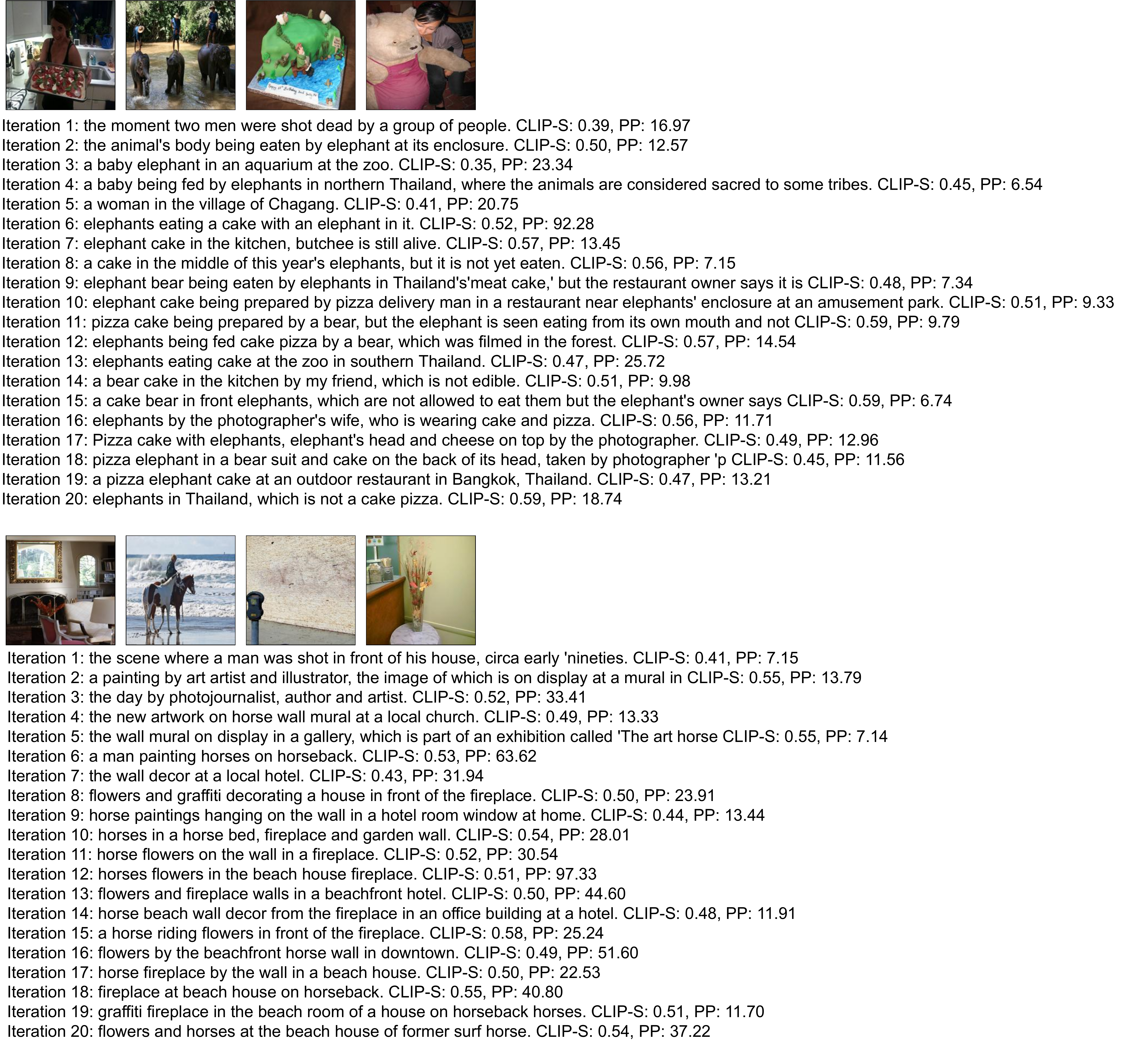}
\end{figure*}

\end{document}